%% file: main.tex
\documentclass[sigplan,10pt]{acmart}

\AtBeginDocument{%
  }

\copyrightyear{2026}
\acmYear{2026}
\setcopyright{cc}
\setcctype{by}
\acmConference[EUROSYS '26]{21st European Conference on Computer Systems}{April 27--30, 2026}{Edinburgh, Scotland Uk}
\acmBooktitle{21st European Conference on Computer Systems (EUROSYS '26), April 27--30, 2026, Edinburgh, Scotland Uk}
\acmPrice{}
\acmDOI{10.1145/3767295.3769319}
\acmISBN{979-8-4007-2212-7/26/04}

\input{header.tex}
\input{_acronyms}

\newcommand{\sys}{\textit{FineMoE}\xspace}
\newcommand{\mixtral}{Mixtral-8$\times$7B\xspace}
\newcommand{\qwen}{Qwen1.5-MoE\xspace}
\newcommand{\phimoe}{Phi-3.5-MoE\xspace}
\newcommand{\lmsys}{LMSYS-Chat-1M\xspace}
\newcommand{\sharegpt}{ShareGPT\xspace}

\begin{document}

\title[\sys: Taming Latency-Memory Trade-Off in MoE-Based LLM Serving]{Taming Latency-Memory Trade-Off in MoE-Based LLM Serving via Fine-Grained Expert Offloading}


\author{\rm Hanfei Yu}
\affiliation{%
  \institution{Stevens Institute of Technology}
  \state{New Jersey}
  \country{USA}
}
\email{hyu42@stevens.edu}

\author{\rm Xingqi Cui}
\authornote{
This work was conducted while Xingqi Cui was a remote intern student, advised by Dr. Hao Wang at the IntelliSys Lab, Stevens Institute of Technology.
}
\affiliation{%
  \institution{Rice University}
  \state{Texas}
  \country{USA}
}
\email{xc66@rice.edu}

\author{\rm Hong Zhang}
\affiliation{%
  \institution{University of Waterloo}
  \state{Ontario}
  \country{Canada}
}
\email{hongzhangblaze@gmail.com}

\author{\rm Hao Wang}
\affiliation{%
  \institution{Rutgers University}
  \state{New Jersey}
  \country{USA}
}
\email{hw488@cs.rutgers.edu}

\author{\rm Hao Wang}
\affiliation{%
  \institution{Stevens Institute of Technology}
  \state{New Jersey}
  \country{USA}
}
\email{hwang9@stevens.edu}

\input{sections/abstract.tex}



\maketitle

\input{sections/intro.tex}

\input{sections/background.tex}

\input{sections/overview.tex}
\input{sections/design.tex}
\input{sections/implement.tex}
\input{sections/eval.tex}

\input{sections/discussion.tex}
\input{sections/related.tex}

\input{sections/conclusion.tex}
\input{sections/acks.tex}

\bibliographystyle{ACM-Reference-Format}
\bibliography{main}

\end{document}

%% file: header.tex
\usepackage{microtype}

\usepackage{bbold}
\usepackage{graphicx}
\usepackage{multirow}
\usepackage{subcaption}
\usepackage{enumitem}
\usepackage{soul}
\usepackage{color}
\usepackage{xcolor}
\usepackage{epsfig}
\usepackage{natbib}
\usepackage{acronym}
\usepackage{pifont}
\usepackage{algorithmic}
\usepackage{tikz}
\usepackage{times}
\usepackage[utf8]{inputenc}
\usepackage{amsmath}
\usepackage{amsfonts}
\usepackage{amsthm}
\usepackage[linesnumbered,ruled,vlined]{algorithm2e}
\usepackage[switch]{lineno}
\usepackage{xspace}
\usepackage{booktabs}
\usepackage{comment}

\usepackage{textcomp}

\usepackage{hyperref}
\hypersetup{
    colorlinks=true,  
    linkcolor=blue,   
    citecolor=red,    
    filecolor=cyan,   
    urlcolor=magenta  
}

\definecolor{ForestGreen}{RGB}{34,139,34}

\usepackage[labelfont=bf]{caption}
\usepackage[T1]{fontenc}
\usepackage{fancyvrb}

\newcommand{\otoprule}{\midrule[\heavyrulewidth]}

\def\hanfei#1{\textcolor{black}{#1}}

\def\revise#1{\textcolor{black}{#1}}

\newcommand{\eg}{{\it e.g.}}

\newcommand{\ie}{{\it i.e.}}
\newcommand{\vs}{\textit{vs.}\xspace}

\newcommand*\circled[1]{\tikz[baseline=(char.base)]{\node[shape=circle,draw,inner sep=1.5pt] (char) {#1};}}

\SetKwSty{text}
\SetArgSty{texttt}
\SetDataSty{}
\SetKwInput{KwInput}{\textcolor{blue}{\textbf{Input}}}
\SetKwInput{KwOutput}{\textcolor{blue}{\textbf{Output}}}
\SetKwIF{If}{ElseIf}{Else}{\textcolor{blue}{\textbf{if}}}{\textcolor{blue}{\textbf{then}}}{\textcolor{blue}{\textbf{else if}}}{\textcolor{blue}{\textbf{else}}}{\textcolor{blue}{\textbf{endif}}}
\SetKwFor{For}{\textcolor{blue}{\textbf{for}}}{\textcolor{blue}{\textbf{do}}}{\textcolor{blue}
{\textbf{endfor}}}
\SetKwFunction{Length}{length}
\SetKwFunction{Max}{max}
\SetKwProg{Fn}{\textcolor{blue}{\textbf{function}}}{}{}
\SetKw{KwRet}{\textcolor{blue}{\textbf{return}}}
\SetKwComment{Comment}{/* }{ */}

%% file: _acronyms.tex
\acrodef{ML}[ML]{Machine Learning}

\acrodef{NSF}[NSF]{National Science Foundation}

\acrodef{AI}[AI]{Artificial Intelligence}

\acrodef{FL}[FL]{Federated Learning}

\acrodef{CL}[CL]{Critical Learning}

\acrodef{AC}[AC]{Attacking-Critical}

\acrodef{CAGR}[CAGR]{compound annual growth rate}

\acrodef{CCT}[CCT]{Center for Computation and Technology}

\acrodef{SLO}[SLO]{service level objective}

\acrodefplural{SLOs}[SLO]{service level objectives}

\acrodef{RL}[RL]{reinforcement learning}

\acrodef{DRL}[DRL]{deep reinforcement learning}

\acrodef{VM}[VM]{virtual machine}

\acrodefplural{VM}[VMs]{virtual machines}

\acrodef{ITC}[ITC]{Innovation \& Technology Commercialization}

\acrodef{DAG}[DAG]{directed acyclic graph}

\acrodefplural{DAG}[DAGs]{directed acyclic graphs}

\acrodef{SFA}[SFA]{single point authentication}

\acrodef{HPC}[HPC]{high-performance computing}

\acrodef{SBIR}[SBIR]{Small Business Innovation Research}

\acrodef{IoT}[IoT]{Internet of Things}

\acrodef{DML}[DML]{distributed machine learning}

\acrodef{GNN}[GNN]{graph neural network}

\acrodefplural{GNN}[GNNs]{graph neural networks}

\acrodef{BSR}[BSR]{backdoor success rate}

\acrodef{BTA}[BTA]{backdoor task accuracy}

\acrodef{ATT}[ATT]{App Tracking Transparency}

\acrodef{DNN}[DNN]{deep neural network}

\acrodef{NN}[NN]{Neural Network}
\newcommand{\NN}{\ac{NN}\xspace}

\acrodefplural{NN}[NNs]{Neural Networks}

\acrodef{KL}[KL]{Kullback–Leibler}

\acrodef{IaaS}[IaaS]{Infrastructure-as-a-Service}

\acrodef{CaaS}[CaaS]{Container-as-a-Service}

\acrodef{TRPO}[TRPO]{Trust Region Policy Optimization}

\acrodef{CPO}[CPO]{Constrained Policy Optimization}

\acrodef{PPO}[PPO]{Proximal Policy Optimization}

\acrodef{TV}[TV]{Total Variation}

\acrodef{PAC}[PAC]{Probably Approximately
Correct}

\acrodef{ACI}[ACI]{Azure Container Instances}

\acrodef{NLP}[NLP]{Natural Language Processing}
\newcommand{\NLP}{\ac{NLP}\xspace}

\acrodef{GAE}[GAE]{Generalized Advantage Estimation}

\acrodef{IS}[IS]{Importance Sampling}

\acrodef{CV}[CV]{coefficient of variance}

\acrodefplural{CV}[CVs]{coefficient of variances}

\acrodef{LLM}[LLM]{Large Language Model}
\newcommand{\LLM}{\ac{LLM}\xspace}

\acrodefplural{LLM}[LLMs]{Large Language Models}
\newcommand{\LLMs}{\acp{LLM}\xspace}

\acrodef{GNS}[GNS]{gradient noise scale}

\acrodef{SOTA}[SOTA]{state-of-the-art}
\newcommand{\SOTA}{\ac{SOTA}\xspace}

\acrodef{SOTP}[SOTP]{state-of-the-practice}

\acrodef{SSP}[SSP]{Stale Synchronous Parallel}

\acrodef{CDF}[CDF]{cumulative distribution function}

\acrodef{PDF}[PDF]{probability density function}

\acrodef{RPC}[RPC]{remote procedure call}

\acrodef{MoE}[MoE]{Mixture-of-Experts}
\newcommand{\MoE}{\ac{MoE}\xspace}

\acrodef{KD}[KD]{knowledge distillation}

\acrodef{DL}[DL]{Deep Learning}
\newcommand{\DL}{\ac{DL}\xspace}

\acrodef{KV}[KV]{key-value}
\newcommand{\KV}{\ac{KV}\xspace}

\acrodef{TTFT}[TTFT]{Time-To-First-Token}
\newcommand{\TTFT}{\ac{TTFT}\xspace}

\acrodef{TPS}[TPS]{Tokens-Per-Second}
\newcommand{\TPS}{\ac{TPS}\xspace}

\acrodef{TPOT}[TPOT]{Time-Per-Output-Token}
\newcommand{\TPOT}{\ac{TPOT}\xspace}

\acrodef{FFN}[FFN]{feed-forward network}
\newcommand{\FFN}{\ac{FFN}\xspace}

\acrodef{LRU}[LRU]{least recently used}
\newcommand{\LRU}{\ac{LRU}\xspace}

\acrodef{LFU}[LFU]{least frequently used}
\newcommand{\LFU}{\ac{LFU}\xspace}

\acrodef{FLOP}[FLOP]{floating point operation}

\acrodefplural{FLOP}[FLOPs]{floating point operations}
\newcommand{\FLOPs}{\acp{FLOP}\xspace}

\acrodef{ILP}[ILP]{integer linear programming}
\newcommand{\ILP}{\ac{ILP}\xspace}

\acrodef{EPM}[EPM]{expert probability map}

\acrodefplural{EPM}[EPMs]{expert probability maps}

\acrodef{FEP}[FEP]{fine-grained expert probability}

\acrodef{EP}[EP]{expert parallelism}
\newcommand{\EP}{\ac{EP}\xspace}

\acrodef{TP}[TP]{tensor parallelism}
\newcommand{\TP}{\ac{TP}\xspace}

\acrodef{DP}[DP]{data parallelism}

%% file: sections/abstract.tex
\begin{abstract}
Large Language Models (LLMs) have gained immense success in revolutionizing various applications, including content generation, search and recommendation, and AI-assisted operations. To reduce high training costs, Mixture-of-Experts (MoE) architecture has become a popular backbone for modern LLMs. However, despite the benefits, serving MoE-based LLMs experience severe memory inefficiency due to sparsely activated experts.
Recent studies propose to offload inactive experts from GPU memory to CPU memory to improve the serving efficiency of MoE models. However, they either incur high inference latency or high model memory footprints due to coarse-grained designs. 

To tame the latency-memory trade-off in MoE serving, we present \sys, a fine-grained expert offloading system for MoE serving that achieves low inference latency with memory efficiency. 
We design \sys to extract fine-grained expert selection patterns from MoE models and semantic hints from input prompts to efficiently guide expert prefetching, caching, and offloading decisions.
\sys is prototyped on top of HuggingFace Transformers and deployed on a six-GPU testbed. Experiments with open-source MoE models and real-world workloads show that \sys reduces inference latency by 47\% and improves expert hit rate by 39\% over state-of-the-art solutions.
\end{abstract}

%% file: sections/intro.tex
\section{Introduction}
\label{sec:intro}

\begin{figure*}[t]
    \centering
    \begin{subfigure}[t]{0.8\textwidth}
        \centering
        \includegraphics[width=.95\linewidth]{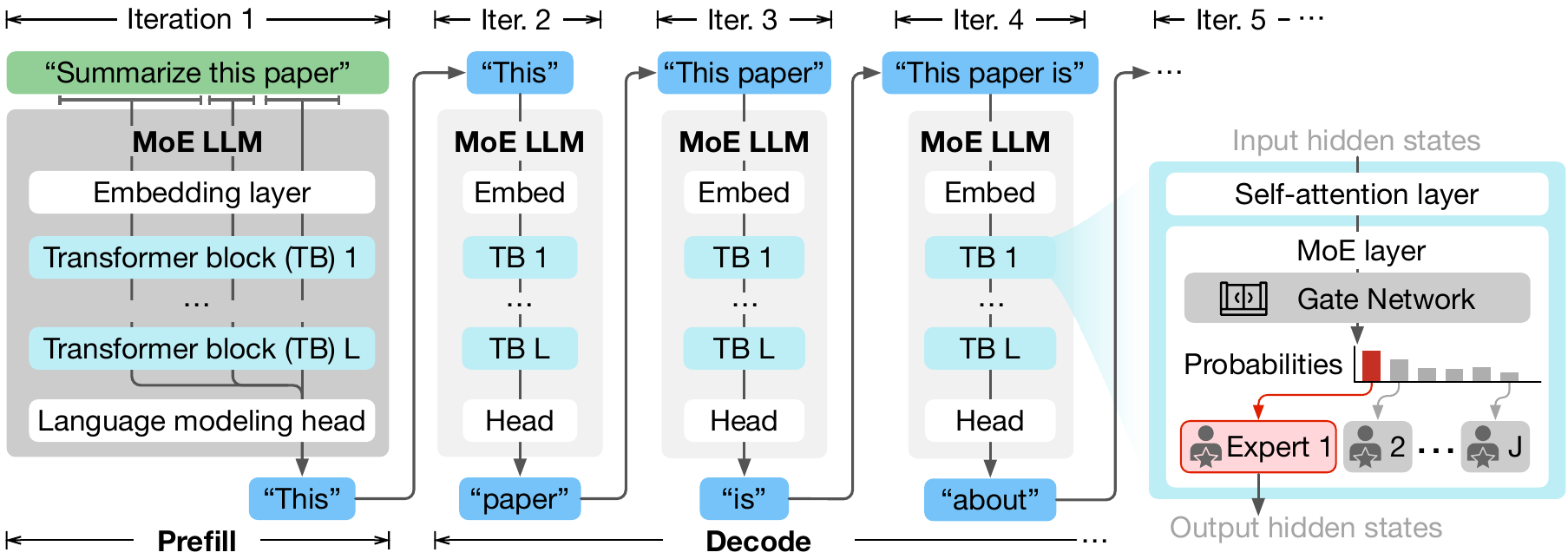}
        \caption{MoE-based LLM serving workflows.}
        \label{fig:bg-moe-workflow}
    \end{subfigure}
    \begin{subfigure}[t]{0.18\textwidth}
        \centering
        \includegraphics[width=\linewidth]{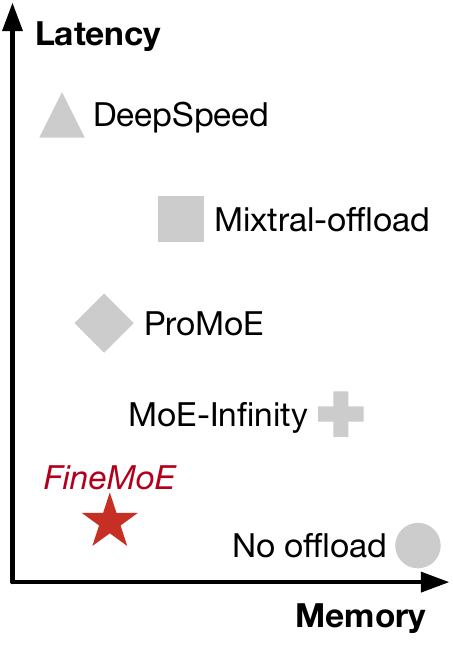}
        \caption{Trade-offs in MoE.}
        \label{fig:bg-trade-off}
    \end{subfigure}
    \caption{Mixture-of-Experts (MoE) Large Language Model (LLM) serving.}
    \label{fig:bg-moe}
\end{figure*}

\LLMs have achieved remarkable success in advancing \NLP research and transforming various applications, including content generation~\cite{dai2024neural,achiam2023gpt,brown2020language,radford2019language}, search and recommendation~\cite{lin2024data,zhao2024let}, and AI-assisted operations~\cite{nam2024using,li2024go,jiang2024lilac}.
Given the high training costs, modern \LLMs have returned to \MoE architectures~\cite{jiang2024mixtral,snowflake-arctic,yang2024qwen2,xai-grok,dai2024deepseekmoe,abdin2024phi} as their backbone implementations. Inside \MoE models, each \MoE layer comprises a gating network and a collection of experts, with only a subset of experts being activated during computation.
This sparse activation mechanism significantly reduces the number of \FLOPs, enabling \MoE-based \LLMs to achieve substantially lower training costs compared to dense \LLMs~\cite{dai2024deepseekmoe,yang2024qwen2,jiang2024mixtral}.

Despite the computational efficiency, \MoE models exhibit substantial memory inefficiency during the serving phase. Though certain model parameters remain inactive during inference, they must still reside in GPU memory to allow for potential future activation.
Expert offloading~\cite{xue2024moe,song2024promoe,eliseev2023fast,aminabadi2022deepspeed} has emerged as a promising strategy to address this issue, which predicts inactive experts and transfers them to CPU memory while retaining only the necessary experts in GPU memory, reducing the overall model memory footprint.

However, existing expert offloading solutions struggle to effectively balance the \textit{latency-memory trade-off} in \MoE serving. These approaches either suffer from high inference latency~\cite{aminabadi2022deepspeed,song2024promoe} or incur substantial model memory footprints~\cite{eliseev2023fast,xue2024moe}.
The key reason is that existing works track expert patterns and manage experts in \textit{coarse granularity}.
They fail to accurately identify and retain only the necessary experts in GPU memory during inference, resulting in frequent and costly on-demand expert loading~\cite{song2024promoe} and performance degradation.

In this paper, we propose \sys, a \textit{fine-grained} expert offloading system that tames the latency-memory trade-off in \MoE serving.
To track and analyze \MoE models' expert selection behaviors in fine granularity, we propose a new data structure called \textit{expert map}, which records the iteration-level probability distributions output by the gate network.
\sys uses historical expert maps for comparing expert trajectory similarity to guide offloading.\footnote{In this paper, ``trajectory'' is defined as the collection of probability distributions over experts observed through layers.}
Apart from the expert map, \sys is designed to track fine-grained input semantic embeddings from individual request prompts processed by the \MoE model.
Given the collected semantic-based and trajectory-based information, \sys carefully searches the most accurate expert map for guiding expert prefetching, caching, and offloading through inference iterations.
In summary, we make the following contributions:
\begin{itemize}[left=10pt,noitemsep,parsep=0pt,partopsep=0pt]
    \item We design \sys, a \textit{fine-grained} expert offloading system that achieves low inference latency while reducing model memory footprints.
    \item We propose a new data structure, expert map, that tracks fine-grained expert selection behaviors of \MoE models. \sys leverages input semantic embeddings to augment the expert map search to guide expert offloading.
    \item We prototype \sys on top of HuggingFace Transformers~\cite{wolf2020huggingface} and deploy it on a six-GPU testbed. Extensive experiments with open-source \MoE models and real-world workloads show that \sys reduces inference latency by 47\% and improves expert hit rate by 39\% compared to state-of-the-art solutions.
\end{itemize}

%% file: sections/background.tex
\section{Background and Motivation}
\label{sec:background}



\subsection{\LLM Serving}


Unlike traditional \DL model inference, Large Language Model (LLM) serving consists of two consecutive stages: \textit{prefill} and \textit{decode}. Figure~\ref{fig:bg-moe-workflow} illustrates the two stages when an \LLM performs inference for a request prompt. 
In the prefill stage, the \LLM first computes the intermediate \KV states of the prompt tokens, prefills the \KV cache~\cite{kwon2023efficient,liu2024cachegen,lee2024infinigen,zhong2024distserve,agrawal2024taming}, and then generates the first answer token. 
In the decode stage, the \LLM sequentially generates the answer to the prompt token-by-token in an auto-regressive manner, where tokens generated previously are used for generating the next token. 

The two stages have their own unique characteristics.  
The prefill stage only requires one \textit{iteration}\footnote{\revise{An iteration refers to a single step in auto-regressive inference that generates one new token. The iteration time denotes the end-to-end latency of this step.}}, processing all tokens in parallel and generating the first answer token. 
The decode stage spans several iterations, generating one token per iteration until the answer is completed. 
Due to the different characteristics of the two stages, recent studies~\cite{patel2024splitwise,zhong2024distserve} have identified that the prefill stage is compute-bounded, while the decode stage is considered memory-bounded. 
Therefore, people typically quantify the serving performance of \LLM two stages using different metrics. 
For the prefill stage, \TTFT is commonly employed, which measures the latency from receiving the user request until generating the first answer token. 
For the decode stage, \TPS or \TPOT is used to measure the generation rate of \LLM serving.

\subsection{\MoE-based \LLM Serving}

\begin{figure}[t]
  \centering
  \includegraphics[width=.95\linewidth]{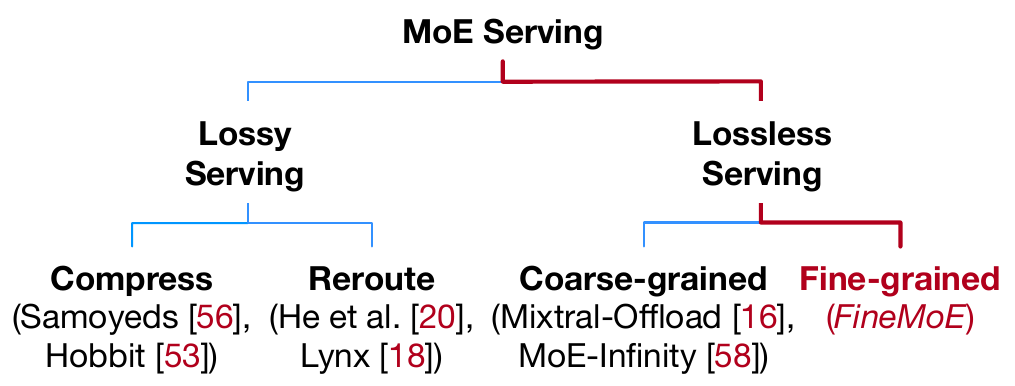}
  \caption{\hanfei{The design space of \MoE-based \LLM serving.}}
  \label{fig:bg-design-space}
\end{figure}


By integrating \MoE layers in Transformer blocks~\cite{vaswani2017attention}, \MoE architectures~\cite{yuksel2012twenty} have emerged as a popular backbone for modern \LLMs, such as Mixtral~\cite{jiang2024mixtral}, Snowflake Arctic~\cite{snowflake-arctic}, and DeepSeek-MoE~\cite{dai2024deepseekmoe}. 
%
%
%
Figure~\ref{fig:bg-moe-workflow} illustrates \MoE-based \LLMs' typical structures, where \FFN modules are replaced by \MoE layers.\footnote{For simplicity, we only show the process of one single request prompt.} 
Each \MoE layer consists of a gate network and a set of expert networks. Inside each Transformer block, the self-attention module first calculates the attentions~\cite{vaswani2017attention} based on input hidden states, and then the gate network determines which expert(s) to activate for computing the output representations. 
Compared to traditional dense \LLMs, \MoE-based \LLMs only activate a subset of parameters during training and inference, reducing computational overhead while delivering superior generation performance compared to dense \LLMs with a comparable number of parameters~\cite{jiang2024mixtral,snowflake-arctic,dai2024deepseekmoe,xai-grok,yang2024qwen2,abdin2024phi}.


Despite the benefits of saving training computations, \MoE-based \LLM serving still suffers from GPU memory inefficiency as \MoE inference requires loading all model parameters into GPU memory, including those inactive experts. 
%
\revise{
Table~\ref{table:moe-models} characterizes three popular \MoE models: \mixtral~\cite{jiang2024mixtral}, \qwen~\cite{yang2024qwen2}, and \phimoe~\cite{abdin2024phi}.
During inference, they exhibit 72\%, 81\%, and 84\% inactive parameters, respectively, due to the sparsity of expert activation in \MoE. 
This corresponds to 67, 23, and 70 GB of inactive GPU memory, resulting in low memory efficiency and serving throughput.
}
Therefore, to efficiently serve large \MoE models, we must seek a solution to the memory inefficiency inherited from \MoE architecture.

\subsection{Latency-Memory Trade-Off}
\label{subsec:bg-latency-memory-tradeoff}






Recently, a few studies have been proposed to improve \MoE-based \LLM serving efficiency.
Figure~\ref{fig:bg-design-space} describes the design space in \MoE serving.
%
%
Existing major studies can be categorized into two types: 
\textbf{Lossy serving} applies compression~\cite{li2023merge}, pruning~\cite{lee2024stun}, and quantization~\cite{kim2023mixture} techniques to the original \MoE models to reduce the serving memory requirements. However, this line of work achieves serving efficiency by sacrificing the generation quality. 
\textbf{Lossless serving} focuses on \textit{offloading} model weights (parameters~\cite{aminabadi2022deepspeed,ollama} or experts~\cite{eliseev2023fast,song2024promoe,xue2024moe}) that are sparsely utilized in temporal or spatial patterns from GPU memory to CPU memory, aiming to preserve reasonable inference latency. 
Specifically, expert offloading seeks to predict the activation of experts in advance, prefetching or caching only the necessary experts in GPU memory during inference.
We opt for lossless serving to design \sys because this line of methods avoids modifying models, hence assuring generation quality. 

However, existing offloading solutions cannot achieve an optimal spot in the latency-memory trade-off when serving \MoE-based \LLMs. 
Figure~\ref{fig:bg-trade-off} compares the performance (\ie, inference latency and memory footprint) of existing \SOTA offloading solutions, which either provide low inference latency but suffer from large memory footprint (\eg, No-offload and MoE-Infinity~\cite{xue2024moe}), or vice versa (\eg, ProMoE~\cite{song2024promoe}, Mixtral-Offloading~\cite{eliseev2023fast}, and DeepSpeed-Inference~\cite{aminabadi2022deepspeed}). 

The key reason behind this dilemma is that \MoE-based decoder-only \LLMs have balanced expert routing~\cite{song2024promoe}, leaving existing solutions hard to find effective patterns for guiding expert offloading.
%
%
Existing research has identified two main reasons for this dilemma:
First, most \MoE-based \LLMs are decoder-only architectures, which exhibit uniform expert activation patterns and low expert access skewness compared to encoder-decoder \MoE \LLMs~\cite{song2024promoe,gupta2024lynx}. 
%
\revise{
Second, recent \MoE-based \LLMs employ a load-balancing loss~\cite{jiang2024mixtral,snowflake-arctic,xai-grok,dai2024deepseekmoe,abdin2024phi}, which encourages the gate network to distribute tokens more uniformly across experts within each \MoE layer, making expert usage more balanced during training.
}
This balanced routing diminishes the predictability of expert patterns, thus making existing solutions ineffective.

\begin{table}[t]
    \centering
    \caption{\revise{Characteristics of three \MoE models.}}
    \scalebox{0.8}{
        \begin{tabular}{lccc}
            \toprule
            \multirow{2}*{\textbf{MoE Models}} & \textbf{Parameters} & \textbf{Experts Per Layer} & \textbf{Num. of} \\
            & \textbf{(active / total)} & \textbf{(active / total)} & \textbf{Layers} \\
            \otoprule 
            \mixtral~\cite{jiang2024mixtral} & 12.9B / 46.7B & 2 / 8 & 32 \\
            \qwen~\cite{yang2024qwen2} & 2.7B / 14.3B & 4 / 60 & 24 \\
            \phimoe~\cite{abdin2024phi} & 6.6B / 42B & 2 / 16 & 32 \\
            \bottomrule 
        \end{tabular}
    }
    \label{table:moe-models}
\end{table}

\subsection{Existing MoE Offloading Solutions}
\label{subsec:bg-coarse-vs-fine}


\begin{figure*}[t]
    \centering
    \begin{subfigure}[t]{0.31\textwidth}
        \centering
        \includegraphics[width=\linewidth]{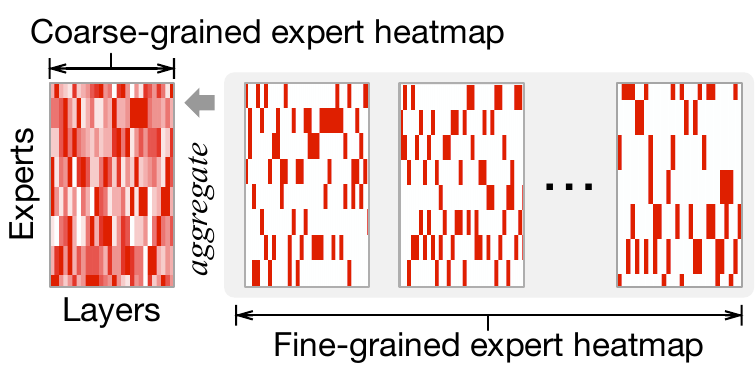}
        \caption{Coarse-grained \vs fine-grained expert heatmaps for \mixtral with \lmsys. Heavier colors indicate more expert activations.}
        \label{fig:bg-expert-heatmap}
    \end{subfigure}
    \hspace{0.02in}
    \begin{subfigure}[t]{0.33\textwidth}
        \centering
        \includegraphics[width=\linewidth]{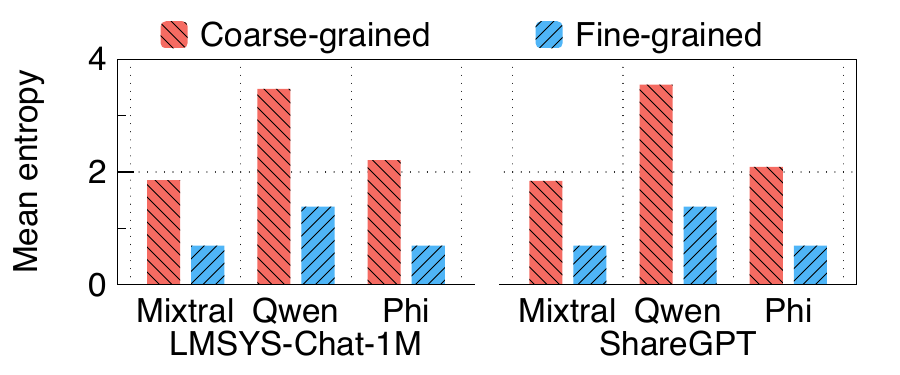}
        \caption{Mean entropy per layer of three MoE models and two datasets for coarse-grained and fine-grained expert patterns. Higher entropy indicates lower predictability.}
        \label{fig:bg-entropy}
    \end{subfigure}
    \hspace{0.02in}
    \begin{subfigure}[t]{0.33\textwidth}
        \centering
        \includegraphics[width=\linewidth]{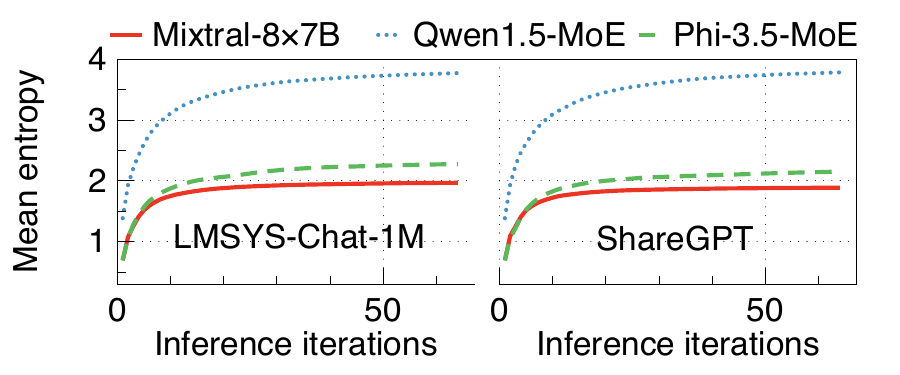}
        \caption{Mean entropy per layer of three MoE models and two datasets when aggregating expert patterns through inference iterations, which diminishes predictability.}
        \label{fig:bg-entropy-iterate}
    \end{subfigure}
    \caption{Expert pattern and predictability analysis in coarse granularity (request-level) and fine granularity (iteration-level).}
    \label{fig:bg-expert-pattern-analysis}
\end{figure*}

Existing expert offloading approaches~\cite{eliseev2023fast,xue2024moe} rely on \textbf{coarse-grained} expert patterns, which are inefficient for guiding offloading. 
%
We define coarse-grained information as the expert patterns collected at the request level, where information is aggregated over multiple iterations of a request prompt.
For example, MoE-Infinity~\cite{xue2024moe} tracks request-level expert activations.
Fine-grained information is defined as the expert patterns observed separately during each inference iteration.
Figure~\ref{fig:bg-expert-heatmap} shows examples of coarse-grained and fine-grained expert activation heatmaps for \mixtral~\cite{jiang2024mixtral}.
The heatmap records the expert activations across 32 \MoE layers, where each layer contains eight experts and activates two experts out of eight to compute representations. 
While fine-grained (iteration-level) heatmaps show clear expert activation patterns, the aggregated coarse-grained (request-level) heatmap diminishes predictability. 

%
To demonstrate this point, we analyze the Shannon entropy~\cite{shannon1948mathematical} of expert activations per \MoE layer for three popular \MoE models.
Entropy is an essential metric to quantify the uncertainty and unpredictability of variables in information theory.
A balanced expert activation pattern (\eg, probability distribution $[0.25, 0.25, 0.25, 0.25]$ of four experts) results in a high entropy, which indicates the pattern is less predictable and harder to select experts.
Figure~\ref{fig:bg-entropy} presents the mean entropy computed per layer for three \MoE models (\mixtral~\cite{jiang2024mixtral}, \qwen~\cite{yang2024qwen2}, and \phimoe~\cite{abdin2024phi}) across two realistic datasets \lmsys~\cite{zheng2023lmsys} and \sharegpt~\cite{sharegpt}.
Coarse-grained expert patterns have significantly higher entropy than fine-grained patterns, meaning that expert patterns in coarse granularity can be less effective for predictions.
\revise{
Figure~\ref{fig:bg-entropy-iterate} shows the mean entropy per layer when aggregating expert patterns across inference iterations, where expert selection becomes increasingly unpredictable as generation progresses. 
\qwen reaches a higher entropy plateau due to its larger expert selection space (60 experts × 24 layers). Similarly, \phimoe (16 × 32) exhibits higher entropy than \mixtral (8 × 32). 
After about ten iterations, expert patterns become blurred and the entropy plateaus, indicating that further iterations contribute only marginal additional unpredictability. 
While entropy is low at the beginning of inference, it gradually increases with iterations as more expert activation information is aggregated, thereby becoming more unpredictable.
}

In contrast to coarse-grained expert offloading solutions, we argue that expert offloading should be carefully guided by \textbf{fine-grained} designs: analyzing iteration-level patterns, understanding models' expert selection preferences, and leveraging semantic characteristics of request prompts. 

\subsection{Problems of Coarse-Grained Offloading}

\begin{figure}[t]
  \centering
  \includegraphics[width=.9\linewidth]{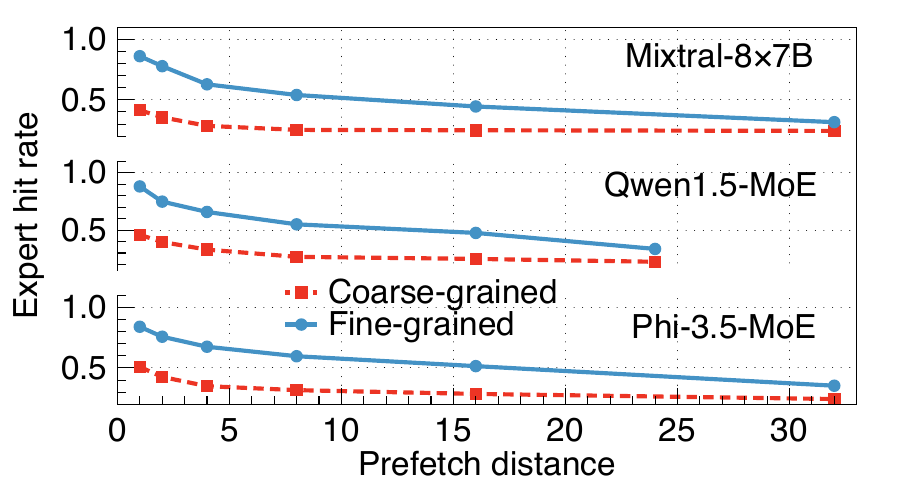}
  \caption{Expert hit rates of coarse-grained and fine-grained expert offloading designs when serving \mixtral, \qwen, and \phimoe with \lmsys at different prefetch distances, respectively.}
  \label{fig:bg-hit-distance}
\end{figure}

Existing coarse-grained expert offloading solutions exhibit three problems:

\noindent \textbf{1) Insufficient latency-memory trade-off.} Existing solutions prefetch and offload experts in coarse granularity, either heavily focusing on reducing inference latency but incurring large memory footprint~\cite{xue2024moe} or reducing memory footprint but severely increasing inference latency~\cite{aminabadi2022deepspeed,eliseev2023fast}.

\noindent \textbf{2) Low expert hit rates.} Existing solutions employ coarse-grained expert pattern tracking methods (\eg, Expert Activation Matrix in MoE-Infinity~\cite{xue2024moe}), which produce ineffective expert patterns for guiding offloading decisions, leading to low expert hit rates and high inference latency.

\noindent \textbf{3) Ignorance of \MoE models' and prompts' heterogeneity.} 
Existing solutions largely ignore the unique characteristics of different \MoE models and input prompts and serve them in a one-fits-all manner~\cite{aminabadi2022deepspeed,eliseev2023fast,song2024promoe,xue2024moe}, which omits opportunities for fine-grained optimizations adaptive to heterogeneous models and prompts in \MoE serving.

%
%
%
Figure~\ref{fig:bg-hit-distance} shows the expert hit rates of serving three popular \MoE-based \LLMs, \mixtral~\cite{jiang2024mixtral}, \qwen~\cite{yang2024qwen2}, and \phimoe~\cite{abdin2024phi} using \lmsys dataset~\cite{zheng2023lmsys} with coarse-grained and fine-grained expert offloading designs at different prefetch distances, respectively. 
Prefetch distance refers to the number of layers ahead that a prefetch instruction is issued before the target layer activates its experts. 
By leveraging fine-grained expert offloading, we can achieve significantly higher expert hit rates over coarse-grained methods and preserve better performance by adapting to varying prefetch distances.

%% file: sections/overview.tex
\section{\sys's Overview}

\subsection{Objectives and Challenges}

\sys is designed to achieve the following three goals: 

\textbf{Memory-efficient \MoE serving with minimal inference latency.}
We have demonstrated that existing expert offloading  solutions~\cite{eliseev2023fast,song2024promoe,xue2024moe} fail to tame the latency-memory trade-off in \MoE serving (\S\ref{subsec:bg-latency-memory-tradeoff}). 
We aim to achieve both low memory footprint and inference latency by proposing fine-grained expert offloading.

\textbf{Minimize expert miss due to mispredictions in expert prefetching.}
Expert prefetching, involving future expert activation predictions, is an essential step in expert offloading solutions. 
However, a recent study~\cite{song2024promoe} has shown that \textit{expert miss} due to mispredictions can cause high on-demand expert loading delay in inference.
We should minimize expert miss and mitigate mispredictions in expert offloading.

\textbf{Adapt to heterogeneous \MoE models and prompts.}
%
\MoE inference can serve heterogeneous models~\cite{jiang2024mixtral,dai2024deepseekmoe,yang2024qwen2,snowflake-arctic,xai-grok} with varying prompts~\cite{zheng2023lmsys,sharegpt} in real-world scenarios.
While existing solutions handle different models and prompts with a one-fits-all design, we should design our expert offloading to adapt to the heterogeneity in \MoE serving.

We must address three critical challenges to realize the above objectives:

\textbf{How to maximize expert hit rate when prefetching and offloading experts?} 
Expert hit rate directly relates to the inference latency. With more experts being hit, fewer experts need to be loaded on demand. 
We propose a fine-grained expert offloading solution to achieve a high expert hit rate.


\textbf{How to adapt to different \MoE models and prompts?}
Heterogeneous \MoE models and input prompts exhibit unique system and semantic characteristics.
We should craft our solution with fine-grained optimizations to enable adaptivity.

\textbf{How to avoid additional system overheads when managing experts?}
Our design must not introduce additional system overheads when serving existing \MoE \LLMs.
We apply a series of system optimizations in \sys to ensure serving efficiency and minimize additional overheads.

\subsection{Architecture and Workflow}

\begin{figure}[t]
  \centering
  \includegraphics[width=.9\linewidth]{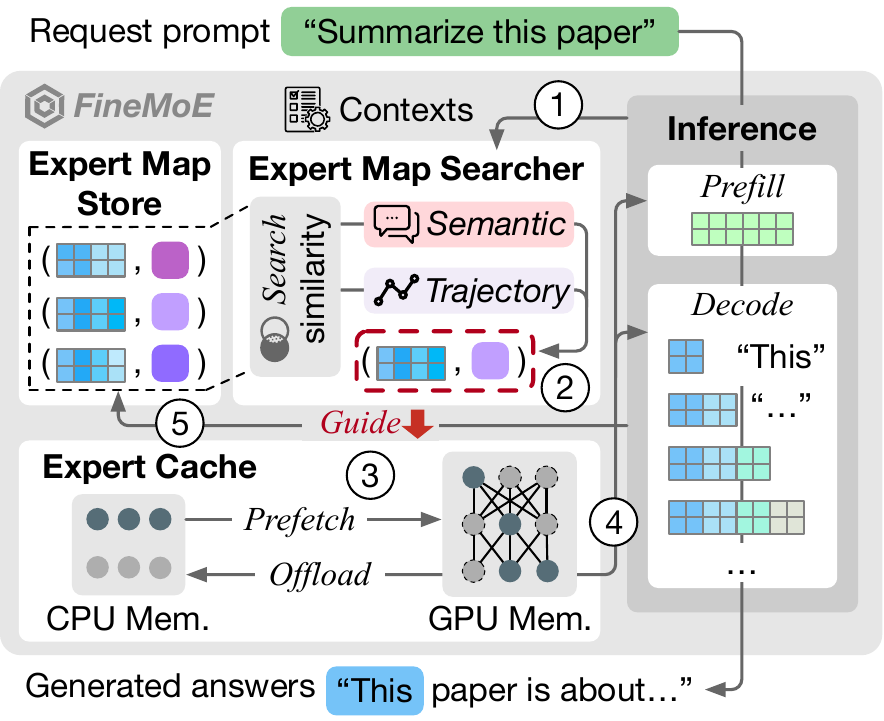}
  \caption{\sys's architecture and workflow.} 
  \label{fig:overview-arch.pdf}
\end{figure}

Figure~\ref{fig:overview-arch.pdf} describes the architecture and workflow of \sys, which consists of three main components: 

\textbf{Expert Map Store.} 
    We record \textit{expert maps}, a new data structure defined in \sys, to track \textit{fine-grained} expert activation patterns from historical request prompts. 
    expert maps provide nuance expert selection preferences over existing coarse-grained expert tracking methods (\eg, Expert Activation Matrix in MoE-Infinity~\cite{xue2024moe}).
    The Expert Map Store dynamically keeps the most useful and unique expert maps for real-time inferences.

\textbf{Expert Map Searcher.} 
    When a request prompt arrives, \sys searches the Expert Map Store for appropriate expert maps to guide expert prefetching before inference. 
    expert map search is guided by calculating similarity scores in two folds: \textit{semantic} and \textit{trajectory} similarity.

\textbf{Expert Cache.} 
    After receiving the searched expert maps, \sys prefetches experts from CPU memory to GPU to perform computations in inference. 
    \sys evicts and offloads low-priority expert weights to CPU memory if exceeding Expert Cache capacity.

\begin{comment}
\begin{itemize}[left=10pt,noitemsep,parsep=0pt,partopsep=0pt]
    \item \textbf{Expert Map Store.} 
    We record \textit{expert maps}, a new data structure defined in \sys, to track \textit{fine-grained} expert activation patterns from historical request prompts. 
    expert maps provide nuance expert selection preferences over existing coarse-grained expert tracking methods (\eg, Expert Activation Matrix in MoE-Infinity~\cite{xue2024moe}).
    The Expert Map Store dynamically keeps the most useful and unique expert maps for real-time inferences.
    %
    \item \textbf{Expert Map Searcher.} 
    When a request prompt arrives, \sys searches the Expert Map Store for appropriate expert maps to guide expert prefetching before inference. 
    expert map search is guided by calculating similarity scores in two folds: \textit{semantic} and \textit{trajectory} similarity.
    %
    \item \textbf{Expert Cache.} 
    After receiving the searched expert maps, \sys prefetches experts from CPU memory to GPU to perform computations in inference. 
    \sys evicts and offloads low-priority expert weights to CPU memory if exceeding Expert Cache capacity.
\end{itemize}
\end{comment}


\sys follows the five steps below to enable memory-efficient \MoE serving with minimal inference latency:

\noindent \textbf{Step {\Large \circled{\small 1}}: Inference context collection.} 
Before every inference iteration, \sys collects necessary \textit{contexts}, such as semantic embeddings and previous expert activation trajectories (\S\ref{subsec:design-expert-map}), and feeds them to the Expert Map Searcher for hybrid similarity searching.

\noindent \textbf{Step {\Large \circled{\small 2}}: Expert map similarity searching.} 
After receiving iteration-level contexts, the Expert Map Searcher identifies the most similar expert maps by comparing the input context data with historical context data in the Expert Map Store (\S\ref{subsec:design-similarity-search}). 
The retrieved expert maps are forwarded to the Expert Cache to guide expert prefetching and offloading decisions.

\noindent \textbf{Step {\Large \circled{\small 3}}: Guided expert prefetching and offloading.} 
We dynamically compute expert selection thresholds to determine which expert(s) to prefetch and offload in the \MoE model guided by the searched expert maps (\S\ref{subsec:design-expert-prefetch}). Then, \sys prefetches the expert weights from CPU to GPU memory and offloads cached experts from GPU to CPU when reaching the cache limit (\S\ref{subsec:design-expert-cache}).

\noindent \textbf{Step {\Large \circled{\small 4}}: Expert serving.} 
The whole inference process consists of one iteration in the Prefill stage and multiple iterations in the Decode stage. For each \MoE layer in every iteration, \sys directly serves the expert required by the gating network if the corresponding weights are available in the GPU memory (defined as an expert hit). Otherwise, \sys on-demand loads the expert weights from CPU to GPU to perform lossless serving (defined as an expert miss).

\noindent \textbf{Step {\Large \circled{\small 5}}: Expert map update.} 
\sys observes new expert maps produced after each iteration and updates them in the Expert Map Store (\S\ref{subsec:design-expert-map-store}). When reaching the store capacity (\eg, 1K expert maps), \sys deduplicates the Expert Map Store by identifying and dropping redundant expert maps to maintain diversity, maximizing the possibility of providing effective expert maps for any request prompts.


\subsection{Problem Formulation}

We consider serving an \MoE-based \LLM with $L$ \MoE layers on a GPU cluster, where each \MoE layer has one gating network and $J$ experts. 
The gating network of each layer selects top $K \in [1, J]$ experts for computation.
The \MoE model processes and generates answers for a workload consisting of $W$ unique request prompts.
\hanfei{
Let $[W] := \{1, \ldots, w, \ldots, W\}$ denote the set of all requests, $[L] := \{1, \ldots, l, \ldots, L\}$ denote the set of all layers in a \MoE model, and $[J] := \{1, \ldots, j, \ldots, J\}$ denote the set of all experts in a layer, respectively.
}
Each request prompt $w \in [W]$ consists of multiple iterations processed during the prefill and decode stages.
Let $E^{(i)}_{l, j}$ denote the $j$-th expert at the $l$-th layer in the $i$-th iteration, where $l \in [L]$, $j \in [J]$, and $i \in [w]$.
During each iteration $i$, we can make at most $L \cdot J$ prefetching decisions. 
Let $E^i_{\textit{cache}}$ and $E^i_{\textit{activate}}$ denote the set of cached experts and the set of activated experts for Iteration $i$, respectively.
\hanfei{
Hence, we represent the result of whether an expert $E^{(i)}_{l, j} \in E^{(i)}_{\textit{activate}}$ is missed by $E^{(i)}_{\textit{cache}}$:
}
\begin{equation*}
  R^{(i)}_{l, j} = 
  \begin{cases}
    1, & \text{if \big{(}$E^{(i)}_{l, j} \in E^{(i)}_{\textit{activate}}\big{)} \bigwedge \big{(}E^{(i)}_{l, j} \notin E^{(i)}_{\textit{cache}}\big{)}$}, \\
    0, & \text{otherwise},
  \end{cases}
\end{equation*}
where $R^{(i)}_{l, j} = 1$ means $E^{(i)}_{l, j}$ is a miss \hanfei{and requires on-demand loading from CPU memory}.
Since all experts in an \MoE model are typically designed to have the same weight size, we assume experts' loading time $T_e$ and memory footprint $M_e$ are homogenous.\footnote{We only consider selective experts. Some MoE models, such as Qwen1.5-MoE-A2.7B, have a few always-on experts that are not offloadable.}
Therefore, the total on-demand loading latency $T$ is summed across all iterations for each expert during the inference process:
\begin{align}
T := T_e \cdot \sum_{w \in [W]} \sum_{i \in [w]} \sum_{l \in [L]} \sum_{j \in [J]} R^{(i)}_{l,j}. \notag
\end{align}
Finally, employing the above definitions, we formulate the \MoE expert offloading as an \ILP optimization problem: 
\begin{align}
    \min_{\{E^{(i)}_{l, j}\}} &\Big{(} T_e \cdot \sum_{w \in [W]} \sum_{i \in [w]} \sum_{l \in [L]} \sum_{j \in [J]} R^i_{l,j} \Big{)} \notag \\
    \mathrm{s.t.~} 
    & |E^{(i)}_{\textit{cache}}| \leq L \cdot J,\quad \forall i \in [w], ~\forall w \in [W], \label{eq:constraint1} \\
    & |E^{(i)}_{\textit{activate}}| = L \cdot K,\quad \forall i \in [w], ~\forall w \in [W], \label{eq:constraint2} \\
    & |E^{(i)}_{\textit{cache}}| \cdot M_e \leq M,\quad \forall i \in [w], ~\forall w \in [W]. \label{eq:constraint3}
\end{align}
The objective is to minimize the on-demand loading latency (ideally $T = 0$ with perfect predictions) while limiting the total memory footprint of cached experts to satisfy the available GPU memory $M$.
Constraint~\ref{eq:constraint1} denotes the total number of prefetched experts should not exceed the total number of all experts in the \MoE model.
Constraint~\ref{eq:constraint2} represents the total number of activated experts, which must be the same as the total number of top $K$ experts summed across all $L$ layers.
Constraint~\ref{eq:constraint3} describes the total memory footprint of prefetched experts must be limited by the available GPU memory size.
Note that solving the \ILP problem is already NP-hard~\cite{cormen2022introduction}, while in reality, prefetching experts always have mispredictions that further complicate the problem. Therefore, we opt for a heuristic-based design for \sys.


%% file: sections/design.tex
\section{\sys's Design}

\begin{figure}[t]
  \centering
  \includegraphics[width=.9\linewidth]{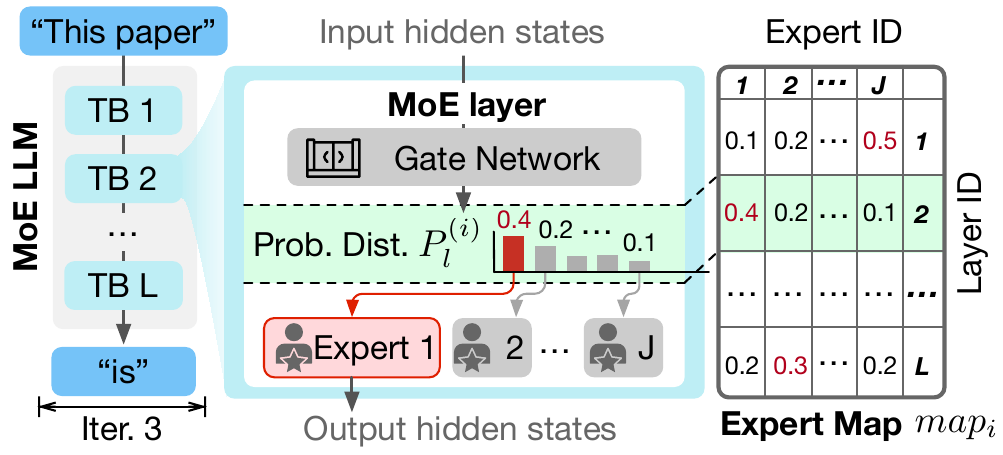}
  \caption{\hanfei{Expert selections tracked by an expert map.}}
  \label{fig:design-expert-map}
\end{figure}

\subsection{Expert Maps}
\label{subsec:design-expert-map}

We propose a new data structure, \textit{Expert Map}, to track expert activation patterns with a fine granularity.
Figure~\ref{fig:design-expert-map} depicts the structure of an expert map. 
During the $i$-th iteration, the $l$-th self-attention layer first calculates the attention states. 
The gate network receives attentions and computes a probability distribution $\mathbf{P}^{(i)}_l \in \mathbb{R}^{J}$ over all the experts at Layer $l$:
\begin{align*}
    \mathbf{P}^{(i)}_l := \big{\{}p^{(i)}_{l,1}, \ldots, p^{(i)}_{l,j}, \ldots, p^{(i)}_{l,J}\big{\}}, \quad \sum_{j \in [J]} p^{(i)}_{l,j} = 1, ~\forall p^{(i)}_{l,j} \geq 0.
\end{align*}
Then, top $K \in [1, J]$ experts are selected from $P^{(i)}_l$ to compute representations for Layer $l$.
We collect the probability distributions $P^{(i)}_l$ across all $L$ layers to form the expert map of Iteration $i$:
\begin{align*}
    \textit{map}_i := \{\mathbf{P}^{(i)}_1, \ldots, \mathbf{P}^{(i)}_l, \ldots, \mathbf{P}^{(i)}_{L}\}, \quad l \in [L].
\end{align*} 

By tracking expert maps, we guide \sys to discover fine-grained expert patterns---the iteration-level expert selection preferences via probability distributions. 
Intuitively, analyzing probability distributions enables \sys to not only identify which experts are binarily selected or omitted, but also to assess the confidence or preference assigned to each expert from the perspective of the gate networks.

The design of expert maps has two key advantages over existing coarse-grained expert tracking methods (\eg, MoE-Infinity~\cite{xue2024moe} tracks the request-level expert hit counts).
\textit{First}, existing works only focus on \textit{aggregated} request-level expert activations, whereas an expert map tracks individual iterations with detailed expert selections.
\textit{Second}, existing works only record the expert hit counts, whereas we track detailed probability distributions. 
Note that expert maps can easily recover coarse-grained information by applying a top $K$ selection operator to the probability distributions and aggregating expert counts over iterations, therefore generalizing to existing tracking methods.
%

\subsection{Expert Map Search}
\label{subsec:design-similarity-search}

\begin{figure}[t]
  \centering
  \includegraphics[width=.95\linewidth]{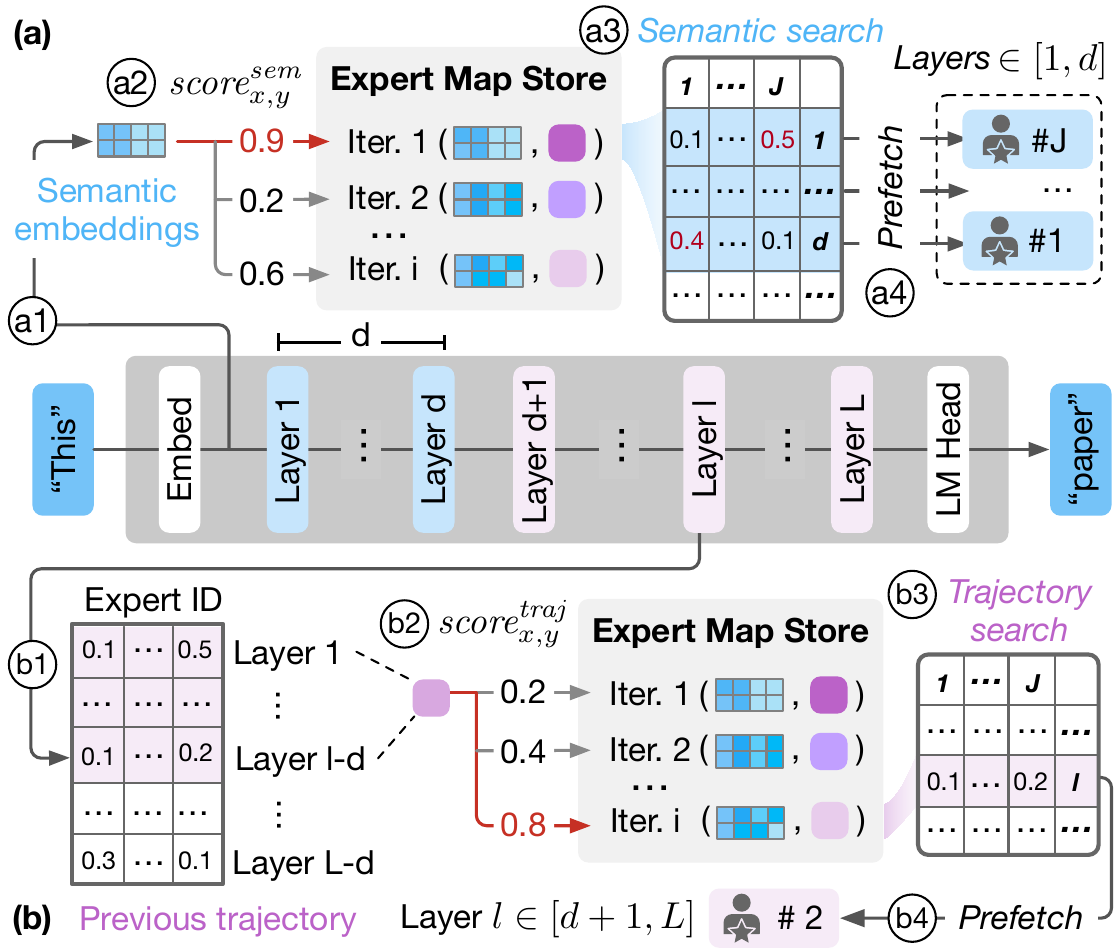}
  \caption{\hanfei{Workflow of \sys's expert map search.}}
  \label{fig:design-map-search}
\end{figure}

\hanfei{
Given the historical expert maps defined in \S\ref{subsec:design-expert-map}, \sys searches expert maps that provide the most accurate expert activation predictions with two fine-grained metrics: semantic similarity (\S\ref{subsubsec:design-semantic-search}) and trajectory similarity (\S\ref{subsubsec:design-trajectory-search}). 
We also show that they are both effective in searching accurate historical expert maps for prediction and offloading (\S\ref{subsubsec:design-similarity-effectiveness}).}

\hanfei{
Existing solutions~\cite{eliseev2023fast,song2024promoe,xue2024moe} \textit{cannot} observe previous expert patterns for prediction and prefetching before the target layer is ready to activate experts for the initial layers $l \in [1, d]$, where $l$ represents the current layer index in an iteration and $d$ is referred as the \textit{prefetch distance}.}
When predicting and prefetching experts for \MoE models, \textit{prefetch distance} is used to avoid impacting inference latency~\cite{song2024promoe,zhang2025daop}.
Prefetch distance is the number of layers ahead that a prefetch instruction is issued before the target layer activates its experts, similar to the same term in memory prefetching~\cite{lee2012prefetching}.
\hanfei{
An ideal prefetch distance should perfectly overlap the prediction and prefetching operation overheads with the inference process.}

Therefore, existing approaches~\cite{eliseev2023fast,song2024promoe,xue2024moe} typically employ coarse-grained rules to prefetch experts for initial layers $l \in [1, d]$. For example, MoE-Infinity~\cite{xue2024moe} prefetches the most popular experts across all historical data points.
Even for layers $l \in [d+1, L]$, existing approaches use coarse-grained (request-level) metrics for predicting and prefetching experts, leading to low offloading accuracy.

\hanfei{
In contrast, \sys leverages fine-grained iteration-level metrics tailored to the prefetch distance $d$, employing semantic embeddings for layers prior to the prefetch distance and expert trajectories for layers subsequent to it.}
Figure~\ref{fig:design-map-search} shows that \sys employs two fine-grained search approaches to jointly search expert maps for guiding expert prefetching:  
%
%
\textit{Semantic-based expert map search} compares the input embeddings with historical embeddings to find expert maps with similar inputs, whereas \textit{trajectory-based search} observes previous expert trajectories (\ie, probability distributions) and searches for similar expert maps. 
We combine both semantic and trajectory features to improve \sys's map-searching and expert offloading accuracy. 
%

\subsubsection{Semantic-based Expert Map Search} 
\label{subsubsec:design-semantic-search}

%
\hanfei{Recent studies~\cite{jie2025mixture} demonstrate that semantic embeddings, \ie, embedding layer's output after processing raw tokens, can potentially indicate expert selection behaviors.}
When serving request prompts and recording their expert maps, we record the \textit{semantic embeddings} for each inference iteration.
%
%
\hanfei{
Existing \MoE-based \LLMs all contain an embedding layer for token semantic encoding, where words or subwords that appear in similar contexts will have similar embeddings~\cite{mikolov2013efficient}. 
It's natural to extract the semantic embeddings using the output from the model's original embedding layer. 
}
\hanfei{
Figure~\ref{fig:design-map-search}(a) shows the semantic-based expert map search in four steps:
a1) extract semantic embeddings from the embedding layer, a2) compute similarity scores using semantic embeddings with historical data points in the Expert Map Store, a3) search similar expert maps based on similarity scores, and a4) prefetch experts with high probabilities for layers $l \in [1, d]$.}

For any input prompts, we compute pairwise cosine similarity $\textit{score}^{\textit{sem}} \in \mathbb{R}^{B \times C}$ between the semantic embedding $\textit{sem}^{\textit{new}} \in \mathbb{R}^{B \times h}$ and the collection of historical semantic embeddings $\textit{sem}^{\textit{old}} \in \mathbb{R}^{C \times h}$ in the Expert Map Store:
\begin{align}
    \textit{score}^{\textit{sem}}_{x,y} := \frac{\textit{sem}^{\textit{new}}_x \cdot \textit{sem}^{\textit{old}}_y}{\|\textit{sem}^{\textit{new}}_x\| \cdot \|\textit{sem}^{\textit{old}}_y\|}, \quad x \in [B], ~y \in [C],
    \label{eq:score-sem}
\end{align}
where $B$ is the batch size of input prompts, $C$ is the Expert Map Store capacity, and $h$ is the hidden dimension size. Then, for prompt $x$, the historical Iteration $y$ with the highest score is selected. 
We use partial expert maps from the selected iteration, $\{\mathbf{P}^{(y)}_1, \ldots, \mathbf{P}^{(y)}_d\} \in \textit{map}^{\textit{old}}_y$, to guide layers $l \in [1, d]$.

\subsubsection{Trajectory-based Expert Map Search}
\label{subsubsec:design-trajectory-search}

We leverage expert probability trajectories of previous $(l-d)$ layers to search expert maps for layers $l \in [d+1, L]$.
\hanfei{Specifically, when $l = d+1$, we use the past expert trajectories from Layer $1$ for prediction; when $l = d+2$, we use the past trajectories from Layers $1$ and $2$; and so on.
When $l = L$ (last layer), we use the past trajectories from Layers $1$ to $L-d$ for prediction.}
\hanfei{
Figure~\ref{fig:design-map-search}(b) shows the trajectory-based expert search for a layer $l \in [d+1, L]$ in four steps:
b1) collect previous trajectory $\{\mathbf{P}_1, \ldots, \mathbf{P}_{l-d}\}$ from Layers $1$ to $l-d$, b2) compute similarity scores using collected trajectories with historical data points in the Expert Map Store, b3) search similar expert maps based on similarity scores, and b4) prefetch experts with high probabilities for the layer $l \in [d+1, L]$.
We repeat this process until the last layer (Layer $L$) is completed.}

Similar to the semantic-based search, we compute pairwise cosine similarity $\textit{score}^{\textit{traj}} \in \mathbb{R}^{B \times C}$ between the observed trajectories, $\textit{map}^{\textit{new}} \in \mathbb{R}^{B \times (l-d)J}$, and the collection of historical expert maps, $\textit{map}^{\textit{old}} \in \mathbb{R}^{C \times (l-d)J}$, in the Expert Map Store: 
\begin{align}
    \textit{score}^{\textit{traj}}_{x,y} := \frac{\textit{map}^{\textit{new}}_x \cdot map^{\textit{old}}_y}{\|\textit{map}^{\textit{new}}_x\| \cdot \|\textit{map}^{\textit{old}}_y\|}, \quad x \in [B], ~y \in [C].
    \label{eq:score-map}
\end{align}
We select the historical iteration with the highest score. Then, we use $\mathbf{P}^{(y)}_{l} \in \textit{map}^{\textit{old}}_y$ from the selected expert map to guide the expert prefetching for the target layer $l \in [d+1, L]$.

\begin{figure}[t]
  \centering
  \includegraphics[width=.9\linewidth]{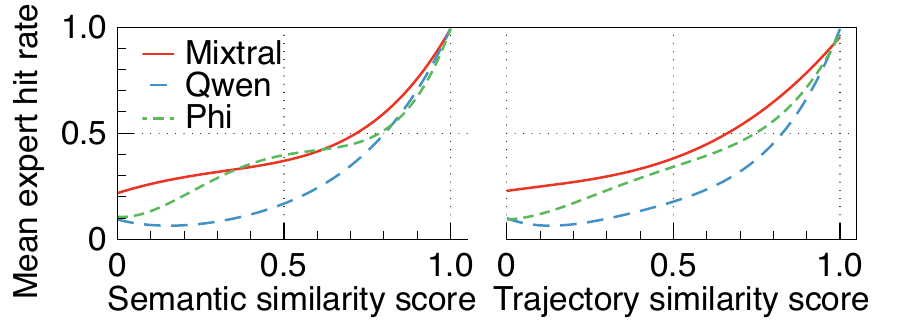}
  \caption{\hanfei{Mean expert hit rates of different semantic and trajectory similarity scores with \lmsys.}}
  \label{fig:design-scores-vs-hit-rates}
\end{figure}

By combining the two expert map search methods, we carefully customize the map that guides expert prefetching for every inference iteration in \MoE serving.
With this design, expert map search introduces negligible overhead to the end-to-end inference latency, which we demonstrate in \S\ref{subsec:eval-overhead}.


\subsubsection{\hanfei{Effectiveness of Semantic and Trajectory Similarity}}
\label{subsubsec:design-similarity-effectiveness}

\hanfei{
To verify how semantic and trajectory similarity scores can guide expert offloading, we run three \MoE models (\mixtral, \qwen, and \phimoe) with two datasets (\lmsys and \sharegpt). 
For each model and dataset, we first run prompts and record their semantic embeddings and expert trajectories, where each prompt generates one data point consisting of a semantic embedding and an expert map. 
Then, we exhaust all pairwise cases by calculating their semantic and trajectory similarity and expert hit rate (\ie, overlapped expert ratio).
Figure~\ref{fig:design-scores-vs-hit-rates} shows the mean expert hit rates of different semantic and trajectory similarity scores for three \MoE models with \lmsys.
Both semantic and trajectory similarity can effectively indicate the accuracy of historical prompts or expert maps for offloading.
}

\begin{figure}[t]
  \centering
  \includegraphics[width=.9\linewidth]{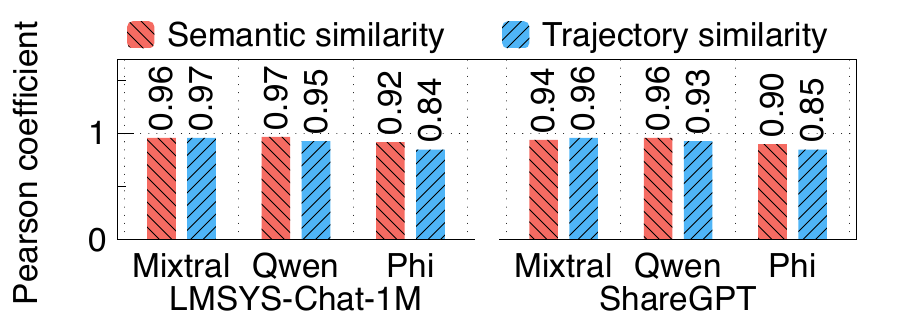}
  \caption{Pearson correlation coefficients between semantic and trajectory similarity scores and expert hit rates.}
  \label{fig:design-correlations}
\end{figure}

\hanfei{
To \hanfei{\textit{statistically}} quantify the correlations between similarity score and expert hit rate, we calculate the Pearson correlation coefficients~\cite{cohen2009pearson} using all paired semantic and trajectory similarity scores and corresponding expert hit rates in Figure~\ref{fig:design-scores-vs-hit-rates}.
The Pearson coefficient is commonly used to measure correlations between variables, where a coefficient close to 1 indicates a strong positive correlation and a coefficient close to 0 means a weak correlation.
Figure~\ref{fig:design-correlations} shows the Pearson coefficients between similarity score and expert hit rate with three \MoE models and two datasets.
The results show that high similarity scores potentially relate to high expert hit rates.
}

\subsection{Expert Prefetching}
\label{subsec:design-expert-prefetch}

\hanfei{
Given the searched and customized expert map $\mathbf{P}^{(i)}_{l}$ for a layer $l \in [L]$ in Iteration $i$, we explain how it guides \sys to dynamically prefetch experts in fine granularity.
}

\textbf{Similarity-aware expert selection.}
With the different contexts collected during iterations, expert maps searched by \sys also have varying similarity scores.\footnote{\hanfei{In the following paper, we use ``similarity scores'' in both search contexts for simplicity, \ie, semantic similarity in semantic-based expert map search and trajectory similarity in trajectory-based search, respectively.}} 
\hanfei{
Figures~\ref{fig:design-scores-vs-hit-rates} and ~\ref{fig:design-correlations} demonstrated that similarity scores can effectively indicate the search confidence, where high searched similarity scores potentially mean high expert hit rates.
}
Hence, we design \sys's expert prefetching to be similarity-aware. 
For a layer $l \in [L]$ with a $\textit{score} \in [-1, 1]$ to prefetch, we first dynamically compute an expert selection threshold $\delta_l \in [0, 1]$:
\begin{align*}
    \delta_l := \textrm{Clip}(1-\textit{score},~0,~1) = \max(0,~\min(1-\textit{score},~1)),
\end{align*}
where $\textit{score}$ is the cosine similarity score computed in Equations~\ref{eq:score-sem} and \ref{eq:score-map}. 
Given searched $\mathbf{P}_l$, we find the set of experts to prefetch $E_{\textit{prefetch}}$ by iteratively picking the expert with the highest probability from $\mathbf{P}_l = \{p_{l,1}, \ldots, p_{l,j}, \ldots, p_{l,J}\}$ until the summed probability of $E_{\textit{prefetch}}$ exceeds $\delta_l$:
\begin{align}
    \min_{\{E_{l,j}\}} &|E_{\textit{prefetch}}| \label{eq:expert-prefetch-set} \\
    \mathrm{s.t.} 
    &\sum_{E_{l,j} \in E_{\textit{prefetch}}} p_{l,j} \geq \delta_l, ~j \in [J], ~\forall l \in [L], \label{eq:expert-prefetch-set-constraint-1} \\
    &|E_{\textit{prefetch}}| \geq K, ~K \leq [J], \label{eq:expert-prefetch-set-constraint-2}
\end{align}
where $K$ is the number of experts needed to activate per layer (\eg, \mixtral activates two experts per layer). 
Constraint~\ref{eq:expert-prefetch-set-constraint-1} requires the total probability of selected experts to prefetch per layer to be greater than $\delta_l$.
Constraint~\ref{eq:expert-prefetch-set-constraint-2} represents the minimum number of selected experts must be larger than the number of experts to activate required by the \MoE model. 
Intuitively, we assign a higher $\delta$ to low-score expert maps so that more experts are prefetched to mitigate mispredictions and assign a lower $\delta$ for high-score expert maps to reduce the memory footprint. 
Experts with higher probabilities are prioritized to be prefetched.

\textbf{Asynchronous expert map searching and prefetching.}
Existing studies~\cite{eliseev2023fast,xue2024moe} predict and prefetch experts synchronously during inference, severely hindering the inference performance. 
For example, MoE-Infinity~\cite{xue2024moe} cannot compute forward functions before finishing expert prediction and prefetching at every \MoE layer~\cite{moe-infinity-code}.
To minimize the system overhead and inference latency, we decouple the map searching and expert prefetching from the inference process using an asynchronous Publisher-Subscriber architecture (Figure~\ref{fig:design-map-search}).
The Expert Map Store is a message broker that keeps messages from both the inference process and the Expert Map Searcher. 
As the inference proceeds, \sys's inference process continuously publishes and writes the inference contexts (\ie, semantic embeddings and expert probability distributions) to the Expert Map Store. At the same time, the Expert Map Searcher subscribes to the context data, searches expert maps based on new context data, and prefetches experts to the Expert Cache in an asynchronous manner.

\subsection{Expert Map Store Management}
\label{subsec:design-expert-map-store}

Practically, we design \sys's Expert Map Store to maintain a capacity $C$ for storing unique expert maps.
To effectively guide inference across diverse prompts, it makes sense to identify and deduplicate redundant expert maps.

\textbf{Expert map deduplication.}
Since \sys uses two approaches (\ie, semantic-based and trajectory-based) to compute similarity, we unify the two similarity scores to compute the pairwise redundancy scores between new iteration data and historical iteration data:
\begin{align*}
    \textit{RDY}_{x,y} := \frac{d}{L} \cdot \textit{score}^{\textit{sem}}_{x,y} + \frac{L-d}{L} \cdot \textit{score}^{\textit{traj}}_{x,y}, \ x \in [B], ~y \in [C],
\end{align*}
where $\textit{score}^{\textit{sem}}_{x,y} \in \mathbb{R}^{B \times C}$ and $\textit{score}^{\textit{traj}}_{x,y} \in \mathbb{R}^{B \times C}$ are semantic-based and trajectory-based pairwise similarity scores calculated from Equations~\ref{eq:score-sem} and \ref{eq:score-map}, $d$ is the prefetch distance, $L$ is the total number of layers, $B$ is the batch size of new interaction data, and $C$ is the Expert Map Store capacity.
Intuitively, as shown in Figure~\ref{fig:design-map-search}, the semantic-based and trajectory-based similarity scores contribute to the search expert map in proportion to $\frac{d}{L}$ and $\frac{L-d}{L}$, respectively. 
Therefore, we follow the same ratio to unify and compute the redundancy score.
Whenever new iterations' context data arrive at the Expert Map Store, we compute the pairwise redundancy score $\textit{RDY}_{x,y}$ to determine which old iterations to drop.
Hence, we update the old iterations $y$ (columns in $\textit{RDY}_{x,y}$) with new iterations $x$ (corresponding rows in $\textit{RDY}_{x,y}$) in the Expert Map Store.

\textbf{Theoretical analysis.}
The expert map deduplication can be formulated as a Minimum Sphere Covering problem~\cite{elzinga1972minimum}.
\hanfei{
Each expert map is a vectorized patch, and the full sphere represents all possible expert selections. 
The objective is to cover as much of the sphere as possible using a small number of maps, keeping storage overhead low.
}
Studies~\cite{rankin1947closest,dumer2007covering} have proved that maintaining at least $2LJ$ expert maps guarantees a lower bound of 75\% expert map similarity (\ie, we can find an expert map that is at least 75\% similar to any new iterations), and keeping $\frac{1}{2}LJ \ln(LJ)$ expert maps provides a lower bound of 98\% similarity, where $L$ and $J$ are the numbers of layers and experts per layer in the \MoE model, respectively.
Given that modern \MoE-based \LLMs generally have $L \in [8, 128]$ and $J \in [24, 96]$, we can approximate the Expert Map Store's maximal requirement to be less than 50K expert maps with 200~MB CPU memory~\cite{xue2024moe}.

\subsection{Expert Caching and Eviction}
\label{subsec:design-expert-cache}

Similar to existing expert offloading solutions~\cite{eliseev2023fast,xue2024moe,song2024promoe}, we design \sys to maintain an Expert Cache on GPUs to reuse expert weights when serving different request prompts.
Given searched expert maps from \S\ref{subsec:design-similarity-search}, we guide \sys's Expert Cache to compute two priority scores for individual experts: 
1) a \textit{prefetching priority} to decide the orders to prefetch experts in the searched maps, and 
2) an \textit{eviction priority} to determine the orders to evict experts in the Expert Cache.

\textbf{Expert prefetching priority.}
Recall the set of experts to prefetch $E_{\textit{prefetch}}$ is determined in Equation~\ref{eq:expert-prefetch-set}. For each expert $E_{l,j} \in E_{\textit{prefetch}}$, we define the prefetching priority to be
\begin{align*}
    PRI^{\textit{prefetch}}_{l, j} := \frac{p_{l,j}}{l-l_{\textit{now}}}, \quad l \in [L], ~j \in [J],
\end{align*}
where $p_{l,j}$ is the expert probability from the searched expert map, and $l_{\textit{now}}$ is the current layer that the inference process stays at.
Intuitively, experts with a higher probability $p_{l,j}$ to be activated should be prefetched sooner, and experts that sit closer to the current layer (\ie, smaller $l - l_{\textit{now}}$) should also be prioritized.

\textbf{Expert eviction priority.}
Similar to MoE-Infinity~\cite{xue2024moe}, \sys's expert caching is based on the \LFU caching algorithm. We integrate the searched map to jointly determine the eviction priority.
For each expert $E_{l,j} \in E_{\textit{cache}}$, we define the eviction priority to be
\begin{align*}
    \textit{PRI}^{\textit{evict}}_{l, j} := \frac{1}{p_{l,j} \cdot \textit{freq}_{l,j}}, \quad l \in [L], ~j \in [J],
\end{align*}
where $\textit{freq}_{l,j}$ is the cache visit frequency and $p_{l,j}$ is the probability from the searched map for an expert $E_{l,j} \in E_{\textit{cache}}$. 
Intuitively, when reaching the Expert Cache limit, we want to first evict experts who are less frequently hit and have lower probabilities of being activated.
Note that similar to existing works~\cite{xue2024moe,song2024promoe}, we do not consider the recent usage of experts as opposed to the classic \LRU algorithm~\cite{eliseev2023fast}.
Since the expert usage is layer-wise sequential, \ie, one layer following another, prioritizing recently used experts is against the nature of sequential forward computation.

\textbf{On-demand expert loading.}
Mispredictions of expert prefetching lead to expert miss in the Expert Cache, as the \MoE model cannot find available experts designated by the gate networks.
Whenever an expert miss occurs, \sys pauses all expert prefetching tasks and immediately loads missed experts from CPU to GPU memory for fast serving.


%% file: sections/implement.tex
\section{\sys's Implementation}
\label{sec:implement}

We prototype \sys on top of Huggingface Transformers framework~\cite{wolf2020huggingface} using MoE-Infinity codebase~\cite{moe-infinity-code}.
The implementation of \sys is described as follows.

\textbf{Expert Map Store} is implemented in Python using PyTorch~\cite{paszke2019pytorch} and NumPy~\cite{harris2020array} libraries.
We store all semantic embeddings and expert maps using \texttt{ndarrays} data structure for efficient array operations. The arrays are converted to tensors to compute similarity for expert map searching.

\textbf{Expert Map Searcher} is implemented in Python using PyTorch~\cite{paszke2019pytorch}.
We implement the pairwise computations, including similarity (\S\ref{subsec:design-similarity-search}) and redundancy (\S~\ref{subsec:design-expert-map-store}) scores, using PyTorch native operations. 

\textbf{Expert Cache} is implemented in C++ based on MoE-Infinity codebase~\cite{moe-infinity-code}. 
The expert management in GPUs is implemented with the CUDA Runtime APIs~\cite{cuda-runtime-api}.
We implement prefetching and caching logic of \sys in the MoE-Infinity codebase to enable expert offloading.
Same with MoE-Infinity, \sys supports multi-GPU inference with \EP, where the experts are mapped to different GPU devices for loading and offloading. 
We use a hash map to assign expert IDs to different GPUs and retrieve them during inference.
The expert assignment follows a round-robin manner to balance the overall GPU load.
Additionally, we use a task pool in the GPU space with asynchronous threads to schedule and execute expert prefetching and on-demand loading tasks.

%% file: sections/eval.tex
\section{Evaluation}

\subsection{Experimental Setup}
\label{subsec:eval-setup}

\hanfei{We introduce our evaluation methodology in this section.}


\textbf{Testbed.}
We conduct all experiments on a six-GPU testbed, where each GPU is an NVIDIA GeForce RTX 3090 with 24 GB GPU memory. 
All GPUs are interconnected using pairwise NVLinks and connected to the CPU memory using PCIe 4.0 with 32GB/s bandwidth. 
Additionally, the testbed has an AMD Ryzen Threadripper PRO 3955WX CPU with 32 cores and 480 GB CPU memory.

\textbf{Models.}
We employ three popular \MoE-based \LLMs in our evaluation: \mixtral~\cite{jiang2024mixtral}, \qwen~\cite{yang2024qwen2}, and \phimoe~\cite{abdin2024phi}.
Table~\ref{table:moe-models} describes the parameters, number of \MoE layers, and number of experts per layer for the three models.
Following the evaluation of existing works~\cite{song2024promoe}, we profile the models to set the optimal prefetch distance $d$ to three before evaluation.

\textbf{Datasets and traces.}
We employ two real-world prompt datasets commonly used for \LLM evaluation: \lmsys~\cite{zheng2023lmsys} and \sharegpt~\cite{sharegpt}.
For most experiments, we split the sampled datasets in a standard 7:3 ratio, where 70\% of the prompts' context data (\ie, semantic embeddings and expert maps) are stored in \sys's Expert Map Store, and 30\% of the prompts are used for testing. 
For online serving experiments, we empty the Expert Map Store and use real-world \LLM inference traces~\cite{patel2024splitwise,stojkovic2025dynamollm} released by Microsoft Azure to set input and generation lengths and drive invocations.

\textbf{Baselines.}
We compare \sys against four \SOTA \MoE serving baselines:
1) \textbf{MoE-Infinity}~\cite{xue2024moe} uses coarse-grained request-level expert activation patterns and synchronous expert prediction and prefetching for \MoE serving. 
We prepare the expert activation matrix collection for MoE-Infinity before evaluation for a fair comparison.
%
2) \textbf{ProMoE}~\cite{song2024promoe} employs a stride-based speculative expert prefetching approach for \MoE serving. Since the codebase of ProMoE is not open-sourced and requires training predictors for each \MoE model, we reproduced a prototype of ProMoE on top of MoE-Infinity in our best effort.
3) \textbf{Mixtral-Offloading}~\cite{eliseev2023fast} combines a layer-wise speculative expert prefetching and a \LRU-based expert cache. 
4) \textbf{DeepSpeend-Inference}~\cite{aminabadi2022deepspeed} employs an expert-agnostic layer-wise parameter offloading approach, which uses pure on-demand loading and does not support prefetching. 
We implement the offloading logic of DeepSpeed-Inference in the MoE-Infinity codebase and add an expert cache for a fair comparison.
We enable all baselines to serve \MoE models from HuggingFace Transformer~\cite{wolf2020huggingface}.

\textbf{Metrics.}
Following the standard evaluation methodology of existing works~\cite{song2024promoe,xue2024moe,zhong2024distserve,agrawal2024taming} on \LLM serving, we report the performance of the prefill and decode stages separately. 
We measure Time-to-First-Token (TTFT) for the prefill stage and Time-Per-Output-Token (TPOT) for the decode stage.
Additionally, we also report other system metrics, such as expert hit rate and overheads, for detailed evaluation.


\subsection{\hanfei{Offline Serving Performance}}
\label{subsec:eval-overall}

\begin{figure}[t]
  \centering
  \includegraphics[width=.9\linewidth]{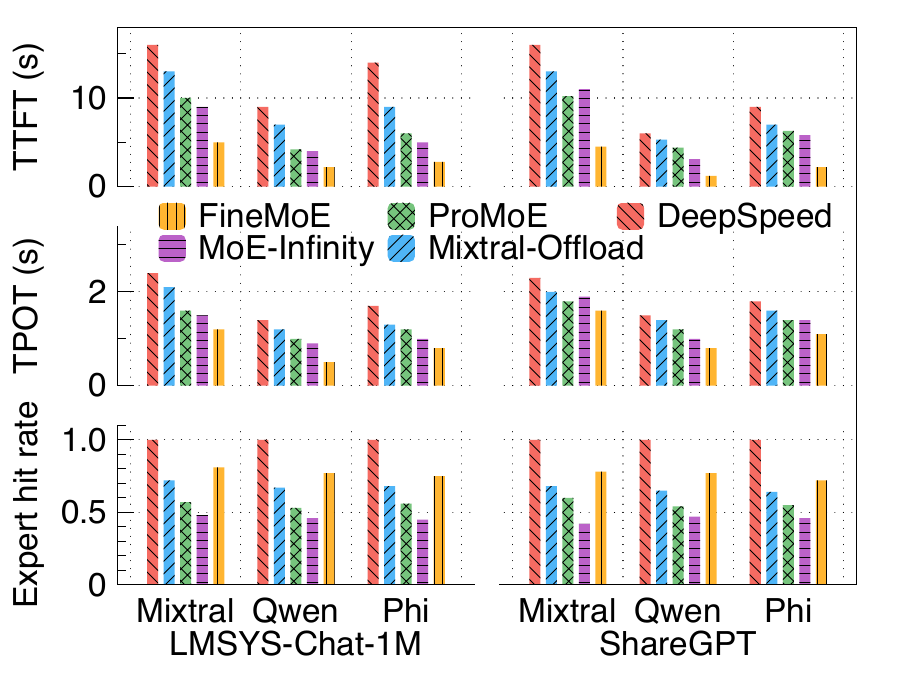}
  \caption{\hanfei{Overall performance of prefill and decode stages.}}
  \label{fig:eval-overall}
\end{figure}

\begin{figure}[t]
  \centering
  \includegraphics[width=.9\linewidth]{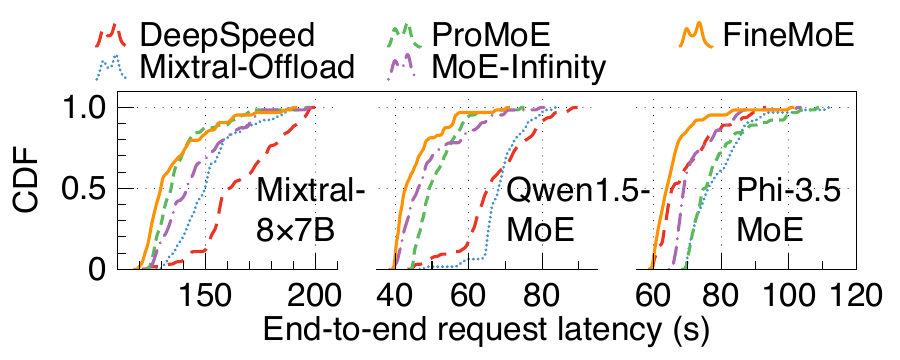}
  \caption{CDF of request latency for \MoE online serving.}
  \label{fig:eval-online-serve}
\end{figure}

We first evaluate the offline serving performance of prefill and decode stages when running \sys and other baselines with the three \MoE models, where we report Time-To-First-Token (TTFT) and Time-Per-Output-Token (TPOT).
\hanfei{
Similar to existing works~\cite{zhong2024distserve,agrawal2024taming}, we measure \TTFT and \TPOT for individual prompts for each combination of model and dataset. For evaluation with \lmsys and \sharegpt datasets, the input lengths are set to 37 and 43 tokens, and generation lengths to 127 and 122 tokens, which are the mean values calculated across datasets, respectively. 
For each dataset, we randomly sample 64 prompts and report average results.}

Figure~\ref{fig:eval-overall} shows the \TTFT, \TPOT, and expert hit rate of \sys and four baselines when serving three \MoE models with \lmsys and \sharegpt datasets, respectively.
DeepSpeed-Inference has both the worst \TTFT and \TPOT due to expert-agnostic offloading and lacking expert prefetching.
While Mixtral-Offloading, ProMoE, and MoE-Infinity perform better than DeepSpeed-Inference, they are underperformed by \sys because of coarse-grained offloading designs.
Compared to DeepSpeed-Inference, Mixtral-Offloading, ProMoE, and MoE-Infinity, \sys reduces the average \TTFT by 74\%, 67\%, 56\%, and 53\%, and reduces the average \TPOT by 46\%, 38\%, 27\%, and 22\%, respectively.

%
\hanfei{
For expert hit rate, DeepSpeed-Inference has no expert misses because it fetches whole layers with full experts, but with the worst latency due to pure on-demand loading.}
Mixtral-Offloading achieves a higher hit rate than ProMoE and MoE-Infinity because of its synchronous speculative prefetching with a prefetch distance of 1. However, due to synchronous prefetching, its \TTFT and \TPOT are worse than others except DeepSpeed-Inference.
Overall, \sys improves the average expert hit rate by 14\%, 37\%, and 68\% over Mixtral-Offloading, ProMoE, and MoE-Infinity, respectively.


\subsection{Online Serving Performance}
\label{subsec:eval-online}

Except for the offline evaluation (\ie, Expert Map Store in full capacity before serving), we also evaluate \sys against other baselines in online serving settings.
We empty the Expert Map Store of \sys and the expert activation matrix collection of MoE-Infinity for the online serving experiment.
The request traces are derived from Azure \LLM inference traces~\cite{patel2024splitwise,stojkovic2025dynamollm}, with randomly sampled 256 requests (2.91 requests per second), to drive \lmsys prompts for each \MoE model serving.
To ensure consistency, \sys and all baselines input and generate the exact number of tokens specified in the traces.
Figure~\ref{fig:eval-online-serve} illustrates the CDF of end-to-end request latency across three \MoE models. The results demonstrate that \sys significantly reduces overall request latency compared to other baselines in online serving.

\begin{figure}[t]
  \centering
  \includegraphics[width=.85\linewidth]{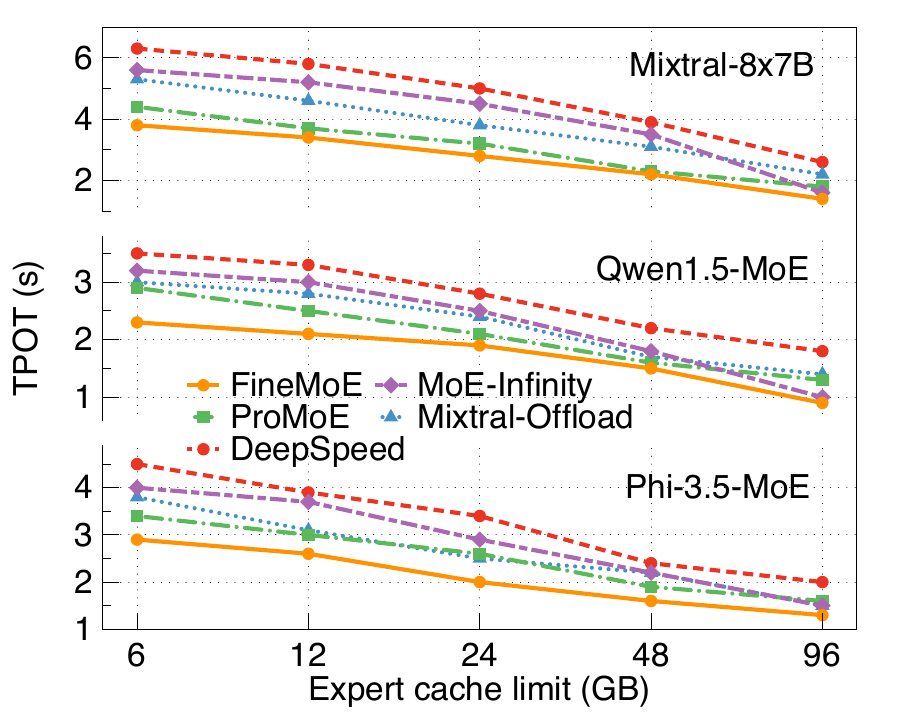}
  \caption{\hanfei{Performance under varying expert cache limits.}}
  \label{fig:eval-cache-limit}
\end{figure}

\subsection{Impact of Expert Cache Limits}

We measure the \TPOT of \sys and other baselines by limiting the expert cache memory budget to investigate their performance in the latency-memory trade-off (\S\ref{subsec:bg-latency-memory-tradeoff}).
We mainly focus on \TPOT to show the end-to-end performance impacted by varying cache limits.
Figure~\ref{fig:eval-cache-limit} shows the \TPOT of \sys and four baselines when serving three \MoE models under different expert cache limits.
We gradually increase the GPU memory allocated for caching experts from 6 GB to 96 GB while employing the same experimental setting in \S\ref{subsec:eval-overall}.
Similarly, DeepSpeed-Inference has the worst \TPOT due to being expert-agnostic.
\sys consistently outperforms Mixtral-Offloading, ProMoE, and MoE-Infinity under varying expert cache limits.
\hanfei{As the cache limit increases, the performance gap between all baselines narrows due to the increased availability of cached experts.}
Nevertheless, for limited GPU memory sizes (\eg, 6GB), \sys reduces the \TPOT by 36\%, 25\%, 16\%, and 29\%, compared to DeepSpeed-Inference, Mixtral-Offloading, ProMoE, and MoE-Infinity, across three \MoE models, respectively.
With fine-grained expert offloading, \sys significantly reduces the expert on-demand loading latency while maintaining a lower GPU memory footprint, therefore achieving a better spot in the latency-memory trade-off of \MoE serving.

\subsection{\revise{Impact of GPU Performance}}

\begin{figure}[t]
  \centering
  \includegraphics[width=.85\linewidth]{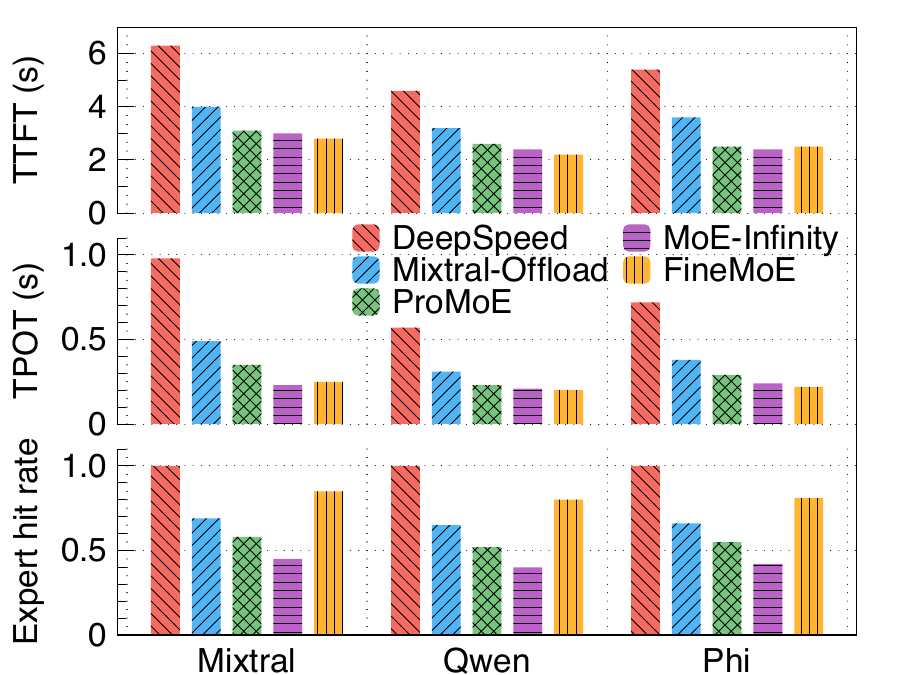}
  \caption{\revise{Performance on high-end GPU testbed.}}
  \label{fig:eval-gpu}
\end{figure}

\revise{
To evaluate the impact of GPU performance on offloading methods, we repeat the experiments using \lmsys on an NVIDIA A100 testbed equipped with 80~GB of HBM2e memory and a peak memory bandwidth of 2~TB/s. 
Figure~\ref{fig:eval-gpu} presents the serving performance of \sys and the baselines across the three \MoE models. 
\sys achieves smaller performance gains on the A100 than on the 6$\times$3090 testbed, since high-end GPUs and the lack of \EP yield faster inference and lower offloading overhead.
Nevertheless, \sys consistently outperforms all baselines. 
The expert hit rate remains largely unaffected, as GPU performance has less impact on expert predictions.
}

\subsection{Ablation Study}
\label{subsec:eval-ablation}

\begin{figure}[t]
    \centering
    \begin{subfigure}[t]{0.595\linewidth}
        \centering
        \includegraphics[width=\linewidth]{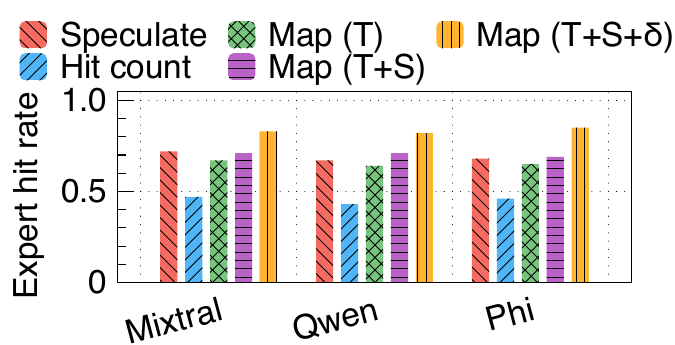}
        \caption{Expert pattern tracking approaches.}
        \label{fig:eval-expert-tracking}
    \end{subfigure}
    \begin{subfigure}[t]{0.395\linewidth}
        \centering
        \includegraphics[width=\linewidth]{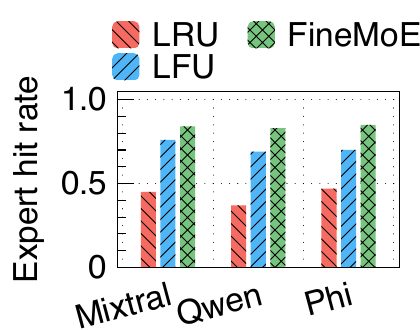}
        \caption{Prefetch and caching.}
        \label{fig:eval-prefetch-and-cache}
    \end{subfigure}
    \vspace{-0.1in}
    \caption{Ablation study of \sys.}
    \label{fig:eval-ablation}
    \vspace{-0.1in}
\end{figure}

We present the ablation study of \sys's design.

\textbf{Effectiveness of expert map search.}
One of \sys's key designs is the expert map, which tracks expert selection preferences in fine granularity.
We evaluate the effectiveness of the expert map against five expert pattern-tracking approaches as follows.
1) \textbf{Speculate}: speculative prediction used by Mixtral-Offloading~\cite{eliseev2023fast} and ProMoE~\cite{song2024promoe}, 
2) \textbf{Hit count}: request-level expert hit count used by MoE-Infinity~\cite{xue2024moe}, 
3) \textbf{Map (T)}: expert map with only trajectory similarity search,
4) \textbf{Map (T+S)}: expert map with both trajectory and semantic similarity search \hanfei{but statically select top-K experts to prefetch},
and
5) \textbf{Map (T+S+$\delta$)}: expert map with full features enabled, including trajectory and semantic similarity search (\S\ref{subsec:design-similarity-search}) and \hanfei{dynamically selecting experts to prefetch} (\S\ref{subsec:design-expert-prefetch}).
We implement the above methods in \sys's Expert Map Searcher for a fair comparison.
Figure~\ref{fig:eval-expert-tracking} shows the expert hit rate of the above expert pattern tracking methods.
Speculative prediction is effective due to the widespread presence of residual connections in Transformer blocks. However, its effectiveness decreases drastically as prefetch distance increases~\cite{song2024promoe}.
The request-level expert activation count has the worst performance due to coarse granularity.
As features are incrementally restored to \sys's expert map, the expert hit rate gradually increases, demonstrating its effectiveness.


\textbf{Effectiveness of expert prefetching and caching.}
We evaluate \sys's expert prefetching and caching against two caching algorithms:
1) \textbf{\LRU} used by Mixtral-Offloading~\cite{eliseev2023fast}
and 
2) \textbf{\LFU} used by MoE-Infinity~\cite{xue2024moe}.
Figure~\ref{fig:eval-prefetch-and-cache} depicts the expert hit rate of \sys and two baselines.
The results show that \LRU performs poorly in expert offloading scenarios. Though \LFU achieves a higher hit rate than \LRU, \sys surpasses both, achieving the highest expert hit rate.

\begin{figure}[t]
  \centering
  \includegraphics[width=.9\linewidth]{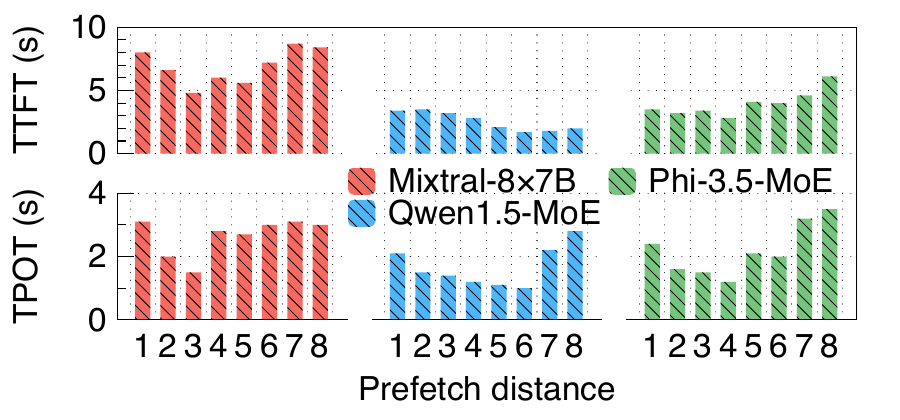}
  \vspace{-0.1in}
  \caption{\hanfei{Performance with different prefetch distances.}}
  \vspace{-0.1in}
  \label{fig:eval-prefetch-distance}
\end{figure}

\begin{figure}[t]
  \centering
  \includegraphics[width=.95\linewidth]{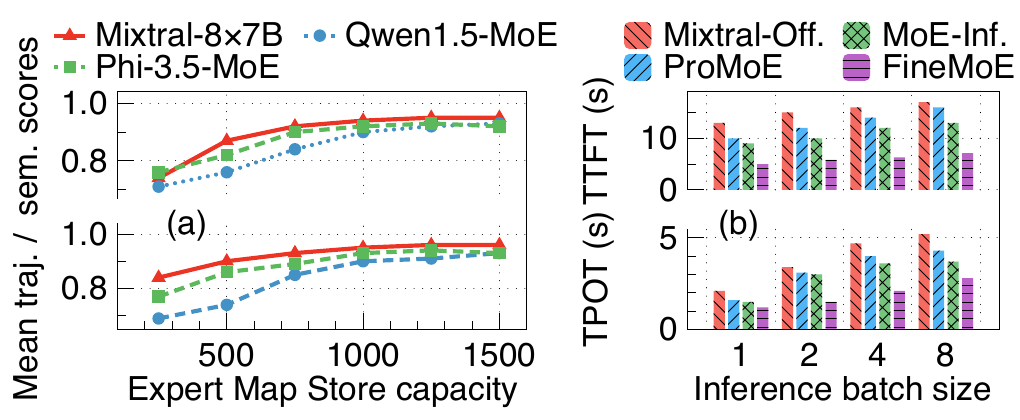}
  \vspace{-0.1in}
  \caption{\hanfei{Sensitivity analysis of \sys.}}
  \vspace{-0.1in}
  \label{fig:eval-sensitivity}
\end{figure}

\subsection{Sensitivity Analysis}
\label{subsec:eval-sensitivity}

We analyze the sensitivity of prefetch distance of \MoE models, Expert Map Store capacity, and inference batch size.

\textbf{Prefetch distance of \MoE models.}
Figure~\ref{fig:eval-prefetch-distance} shows the \TTFT and \TPOT of \sys when serving three \MoE models with different prefetch distances.
We have demonstrated that the expert hit rate decreases when gradually increasing the prefetch distance (Figure~\ref{fig:bg-hit-distance}).
When the prefetch distance is small, \sys cannot perfectly hide its system delay from the inference process, such as the map searching and expert prefetching, leading to an increase in inference latency.
With larger prefetch distances, \sys has worse expert hit rates that also degrade performance. 
Therefore, we set the prefetch distance $d$ to 3, 6, and 4 for \mixtral, \qwen, and \phimoe, respectively.

\textbf{Capacity of Expert Map Store.}
We measure the mean semantic and trajectory similarity scores searched in \sys's expert map searching for \MoE model serving.
Figure~\ref{fig:eval-sensitivity}(a) presents the mean semantic and trajectory similarity scores of \sys with different Expert Map Store capacity sizes.
Both semantic and trajectory similarity scores improve as the store capacity increases.
While the similarity scores exhibit a significant increase with capacities below 1K, further capacity expansion yields diminishing similarity gains. 
To minimize \sys's memory overhead, we set \sys's Expert Map Store capacity to 1K in evaluation.

\textbf{Inference batch size.}
We investigate the impact of inference batch size on \sys and three baselines using \mixtral with \lmsys.
Figure~\ref{fig:eval-sensitivity}(b) presents the performance of \sys, Mixtral-Offloading, ProMoE, and MoE-Infinity as the batch size increases from one to eight. \sys achieves the lowest \TTFT and \TPOT in most cases.

\begin{figure}[t]
  \centering
  \includegraphics[width=.9\linewidth]{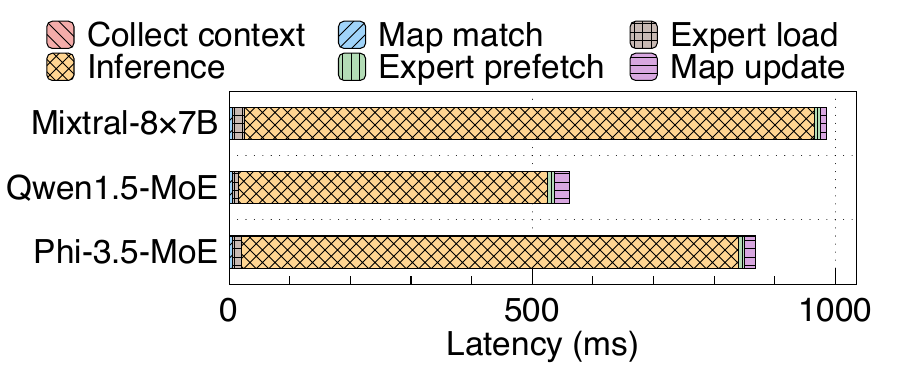}
  \vspace{-0.1in}
  \caption{\hanfei{Latency breakdown of \sys's one iteration.}}
  \vspace{-0.1in}
  \label{fig:eval-overhead-latency.pdf}
\end{figure}

\subsection{System Overheads}
\label{subsec:eval-overhead}

\hanfei{
We measure and report the system overheads of \sys.}

\textbf{Latency overheads of \sys's operations.}
Figure~\ref{fig:eval-overhead-latency.pdf} shows the latency breakdown of one inference iteration in \sys when serving the three \MoE models.
We report operation overheads of \sys, including context collection, map searching, expert on-demand loading, expert prefetching, and map update after the iteration completes.
\qwen has lower end-to-end iteration latency than \mixtral and \phimoe because of significantly fewer parameters.
Note that expert prefetching, map searching, and map update tasks are executed asynchronously, aside from the inference process. Hence, they do not contribute to the end-to-end iteration latency.
Excluding three asynchronous tasks, the total delay incurred by other operations is consistently less than 50ms (1\% of the iteration) across three \MoE models, which is negligible compared to the inference latency.

\begin{figure}[t]
  \centering
  \includegraphics[width=.9\linewidth]{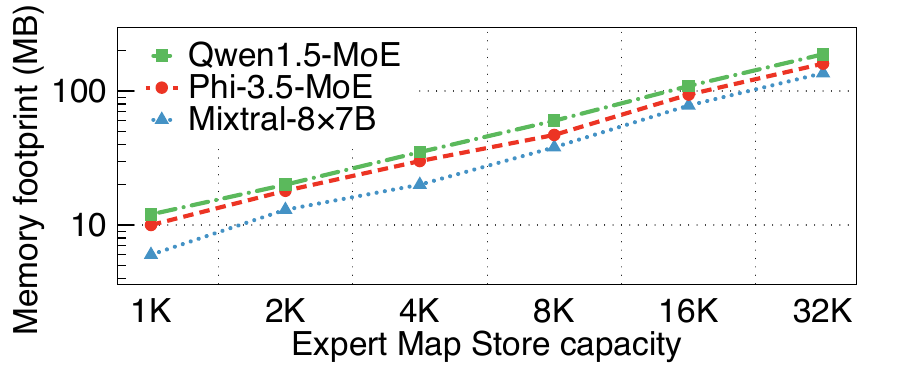}
  \vspace{-0.1in}
  \caption{CPU memory footprint of \sys's Expert Map Store with different capacity.}
  \vspace{-0.1in}
  \label{fig:eval-overhead-memory.pdf}
\end{figure}

\textbf{Memory overheads of \sys's Expert Map Store.}
Figure~\ref{fig:eval-overhead-memory.pdf} shows the CPU memory footprint of \sys's Expert Map Store when varying the store capacity from 1K to 32K maps.
The memory needed to store expert maps for \qwen is more than \mixtral and \phimoe because it has more experts per layer over the other two models, which increases the map shape.
Even for the largest capacity (32K), the Expert Map Store requires less than 200MB of memory to store the maps, which is trivial since modern GPU servers usually have abundant CPU memory (\eg, \texttt{p4d.24xlarge} on AWS EC2~\cite{aws-ec2} has over 1100 GB of CPU memory).
In evaluation, \sys's map store capacity with 1K maps is sufficient for maintaining performance (\S\ref{subsec:eval-sensitivity}), resulting in minimal memory overhead.

%% file: sections/discussion.tex
\section{\revise{Discussion}}
\label{sec:discuss}

\revise{
In this section, we compare the heuristic-based \sys with \NN-based predictors, analyze the impact of model parallelism on \sys's performance, and discuss how \sys can be extended to other \MoE architectures.
}

\revise{
\textbf{\NN-based predictors.}
\NN-based predictors for expert offloading are impractical due to multiple sources of overhead. First, they often introduce sub-second inference latency, comparable to \MoE inference latency itself.
Second, they require extensive data collection, hour-long per-layer training, and frequent retraining to adapt to workload shifts. 
Third, they consume substantial GPU memory, as prior work~\cite{song2024promoe} reports millions of parameters per \MoE layer. 
Moreover, they are incompatible with \sys’s fine-grained design: training on fine-grained data hinders convergence, while storing iteration-level probabilities generates large volumes that further prolong training and limit feasibility. 
Therefore, we adopt a heuristic-based design rather than \NN-based approaches.
}

\revise{
\textbf{Impact of \EP and \TP.}
Higher \EP distributes experts across more devices and enables greater expert replication, which can increase \sys’s offloading opportunities and memory savings. 
In contrast, higher \TP raises the overhead of offloading operations, since dense model components are split across devices and require coordinated offloading and reloading. 
As noted in prior work~\cite{liu2024deepseek,dai2024deepseekmoe}, production \MoE systems generally avoid high \TP because its communication costs outweigh performance benefits. 
In large-scale deployments (\eg, DeepSeek~\cite{liu2024deepseek}), \MoE systems usually use low \EP during prefill to maximize throughput, while adopting high \EP during decode to enable higher expert redundancy. 
Though high-\EP decode reduces per-GPU expert occupancy, the larger number of expert replicas (\eg, 2$\times$ more than prefill in DeepSeek~\cite{liu2024deepseek}) creates additional offloading opportunities by allowing experts to be compacted onto fewer devices.
}

\revise{
\textbf{Adaptation to other \MoE architectures..}
\sys can be easily integrated with different \MoE architectures. 
For shared experts, we treat them as always-hit during expert prediction. One of our evaluated models, \qwen, includes shared experts that are used by all tokens. 
For multi-gating \MoE, we can extend the expert map search by recording each gate’s probability distribution and flattening the outputs into a single vector for efficient similarity computation. 
This enables unified handling across diverse routing schemes.
}

%% file: sections/related.tex
\section{Related Work}

In this section, we provide a brief overview of recent studies and related works on \MoE serving.


\textbf{Lossless \MoE serving.}
Recent studies on lossless \MoE serving have been widely proposed. 
DeepSpeed Inference~\cite{aminabadi2022deepspeed} offloads layer-wise parameters without considering expert awareness and does not provide expert prefetching or caching capabilities.
Mixtral-Offloading~\cite{eliseev2023fast} employs \LRU expert caching and introduces speculative prediction to enable expert prefetching. 
MoE-Infinity~\cite{xue2024moe} proposes the request-level expert activation matrix to guide offloading in coarse granularity.
SwapMoE~\cite{swapmoe} maintains a set of critical experts in GPU memory and adjusts them based on workload changes to minimize offloading overhead.
ProMoE~\cite{song2024promoe} trains predictors per \MoE layer to achieve high speculative prediction accuracy and low inference latency. 
Lina~\cite{li2023accelerating} exports unpopular experts to host memory while focusing on \MoE training. 
\citet{liu2025optimizing} partitions and serves \MoE models on serverless computing.
Fiddler~\cite{kamahori2025fiddler} serves \MoE inference on CPU and GPU collaboratively.
MoEShard~\cite{balmau2025accelerating} shards experts to achieve balanced expert loads.
Unlike existing coarse-grained offloading solutions, \sys tracks fine-grained expert patterns from both trajectory and semantic aspects and outperforms \SOTA baselines.

\textbf{Lossy \MoE serving.}
Expert pruning~\cite{kim2021moe} reduces memory usage by removing under-utilized experts. 
Expert compression~\cite{li2023merge,tang2024hobbit,zhou2025floe,wu2025samoyeds,he2024towards} compresses less-popular experts to reduce models' memory footprint. 
Expert load rerouting~\cite{gupta2024lynx,zhang2025daop,he2025capacity} balances tokens to under-loaded experts instead of following the outputs of gate networks.
Specifically, Hobbit~\cite{tang2024hobbit} uses low precision to serve less-critical experts.
\citet{he2025capacity} drops and reroutes tokens from overloaded experts to others to reduce the straggler effect.
Samoyeds~\cite{wu2025samoyeds} serves \MoE models with sparsity computing.
Lynx~\cite{gupta2024lynx} selects experts based on batch-level expert importance instead of gate network outputs.
DAOP~\cite{zhang2025daop} performs computations with predicted experts directly and cannot guarantee full generation quality.
FLoE~\cite{zhou2025floe} compresses experts on-the-fly during inference.
However, lossy serving impacts the model quality and is orthogonal to \sys.

\textbf{\MoE refactorization.}
Some works propose to redesign and refactor the current \MoE architecture, \hanfei{such as decoupling gate networks from inference process~\cite{hwang2024pre,du2024sida} or building activation-efficient \MoE models~\cite{cai2024read,jie2025mixture}.
This line of work requires model training or fine-tuning before serving.}
Pre-gated MoE~\cite{hwang2024pre} trains pre-gate functions to eliminate the sequential dependencies between expert selection and execution. 
SiDA~\cite{du2024sida} proposes a sparsity-inspired data-aware inference system that decouples the expert routing from inference. 
READ-ME~\cite{cai2024read} refactors pre-trained dense \LLMs into specialized \MoE models.
\hanfei{
MoLE~\cite{jie2025mixture} replaces inputs of all \MoE layers with embedding tokens to avoid sparse expert activation.
}
In contrast, \sys requires zero training to serve \MoE models.

%% file: sections/conclusion.tex
\section{Conclusion}

This paper proposes \sys, a fine-grained expert offloading system for \MoE serving that achieves low inference latency without incurring significant model memory footprints.
\sys tracks iteration-level expert probability distributions from the \MoE model using expert map and analyzes input semantic embeddings from individual request prompts.
Based on the input semantic and expert trajectory information, \sys searches the most accurate expert map to carefully guide the expert prefetching, caching, and offloading decisions tailored to every inference iteration.
\sys is prototyped on top of HugginFace Transformers and deployed to a six-GPU testbed.
Extensive experiments with open-source \MoE models and real-world workloads show that \sys reduces inference latency by 47\% and improves expert hit rate by 39\% compared to state-of-the-art solutions.

%% file: sections/acks.tex
\begin{acks}
We thank anonymous reviewers and our shepherd, Dr. Yaniv David, for their valuable feedback.
The work of Hanfei Yu and Hao Wang was supported in part by NSF 2527416, 2534241, and 2523997, and the AWS Cloud Credit for Research program. 
%
%
The work of Hao Wang (Rutgers CS) was supported in part by Amazon Faculty Research Award, Microsoft AI \& Society Fellowship, NSF CAREER Award IIS-2340125, NIH grant R01CA297832, and NSF grant IIS-2127918. 
Any opinions, findings, and conclusions or recommendations expressed in this material are those of the authors and do not necessarily reflect the views of the funding agencies.
\end{acks}

%% file: main.bbl

\begin{thebibliography}{66}


\ifx \showCODEN    \undefined \def \showCODEN     #1{\unskip}     \fi
\ifx \showISBNx    \undefined \def \showISBNx     #1{\unskip}     \fi
\ifx \showISBNxiii \undefined \def \showISBNxiii  #1{\unskip}     \fi
\ifx \showISSN     \undefined \def \showISSN      #1{\unskip}     \fi
\ifx \showLCCN     \undefined \def \showLCCN      #1{\unskip}     \fi
\ifx \shownote     \undefined \def \shownote      #1{#1}          \fi
\ifx \showarticletitle \undefined \def \showarticletitle #1{#1}   \fi
\ifx \showURL      \undefined \def \showURL       {\relax}        \fi
\providecommand\bibfield[2]{#2}
\providecommand\bibinfo[2]{#2}
\providecommand\natexlab[1]{#1}
\providecommand\showeprint[2][]{arXiv:#2}

\bibitem[Abdin et~al\mbox{.}(2024)]%
        {abdin2024phi}
\bibfield{author}{\bibinfo{person}{Marah Abdin}, \bibinfo{person}{Jyoti Aneja}, \bibinfo{person}{Hany Awadalla}, \bibinfo{person}{Ahmed Awadallah}, \bibinfo{person}{Ammar~Ahmad Awan}, \bibinfo{person}{Nguyen Bach}, \bibinfo{person}{Amit Bahree}, \bibinfo{person}{Arash Bakhtiari}, \bibinfo{person}{Jianmin Bao}, \bibinfo{person}{Harkirat Behl}, {et~al\mbox{.}}} \bibinfo{year}{2024}\natexlab{}.
\newblock \showarticletitle{{Phi-3 Technical Report: A Highly Capable Language Model Locally on Your Phone}}.
\newblock \bibinfo{journal}{\emph{arXiv preprint arXiv:2404.14219}} (\bibinfo{year}{2024}).
\newblock


\bibitem[Achiam et~al\mbox{.}(2023)]%
        {achiam2023gpt}
\bibfield{author}{\bibinfo{person}{Josh Achiam}, \bibinfo{person}{Steven Adler}, \bibinfo{person}{Sandhini Agarwal}, \bibinfo{person}{Lama Ahmad}, \bibinfo{person}{Ilge Akkaya}, \bibinfo{person}{Florencia~Leoni Aleman}, \bibinfo{person}{Diogo Almeida}, \bibinfo{person}{Janko Altenschmidt}, \bibinfo{person}{Sam Altman}, \bibinfo{person}{Shyamal Anadkat}, {et~al\mbox{.}}} \bibinfo{year}{2023}\natexlab{}.
\newblock \showarticletitle{{GPT-4 Technical Report}}.
\newblock \bibinfo{journal}{\emph{arXiv preprint arXiv:2303.08774}} (\bibinfo{year}{2023}).
\newblock


\bibitem[Agrawal et~al\mbox{.}(2024)]%
        {agrawal2024taming}
\bibfield{author}{\bibinfo{person}{Amey Agrawal}, \bibinfo{person}{Nitin Kedia}, \bibinfo{person}{Ashish Panwar}, \bibinfo{person}{Jayashree Mohan}, \bibinfo{person}{Nipun Kwatra}, \bibinfo{person}{Bhargav Gulavani}, \bibinfo{person}{Alexey Tumanov}, {and} \bibinfo{person}{Ramachandran Ramjee}.} \bibinfo{year}{2024}\natexlab{}.
\newblock \showarticletitle{{Taming Throughput-Latency Tradeoff in LLM Inference with Sarathi-Serve}}. In \bibinfo{booktitle}{\emph{18th USENIX Symposium on Operating Systems Design and Implementation (OSDI)}}.
\newblock


\bibitem[Aminabadi et~al\mbox{.}(2022)]%
        {aminabadi2022deepspeed}
\bibfield{author}{\bibinfo{person}{Reza~Yazdani Aminabadi}, \bibinfo{person}{Samyam Rajbhandari}, \bibinfo{person}{Ammar~Ahmad Awan}, \bibinfo{person}{Cheng Li}, \bibinfo{person}{Du Li}, \bibinfo{person}{Elton Zheng}, \bibinfo{person}{Olatunji Ruwase}, \bibinfo{person}{Shaden Smith}, \bibinfo{person}{Minjia Zhang}, \bibinfo{person}{Jeff Rasley}, {et~al\mbox{.}}} \bibinfo{year}{2022}\natexlab{}.
\newblock \showarticletitle{{DeepSpeed-Inference: Enabling Efficient Inference of Transformer Models at Unprecedented Scale}}. In \bibinfo{booktitle}{\emph{SC22: International Conference for High Performance Computing, Networking, Storage and Analysis}}.
\newblock


\bibitem[{AWS}(2006)]%
        {aws-ec2}
\bibfield{author}{\bibinfo{person}{{AWS}}.} \bibinfo{year}{2006}\natexlab{}.
\newblock \bibinfo{title}{{AWS EC2: Secure and Resizable Compute Capacity in the Cloud}}.
\newblock \bibinfo{howpublished}{\url{https://aws.amazon.com/ec2/}}.
\newblock


\bibitem[Balmau et~al\mbox{.}(2025)]%
        {balmau2025accelerating}
\bibfield{author}{\bibinfo{person}{Oana Balmau}, \bibinfo{person}{Anne-Marie Kermarrec}, \bibinfo{person}{Rafael Pires}, \bibinfo{person}{Andr{\'e} Loureiro~Esp{\'\i}rito Santo}, \bibinfo{person}{Martijn de Vos}, {and} \bibinfo{person}{Milos Vujasinovic}.} \bibinfo{year}{2025}\natexlab{}.
\newblock \showarticletitle{{Accelerating MoE Model Inference with Expert Sharding}}. In \bibinfo{booktitle}{\emph{Proceedings of the 5th Workshop on Machine Learning and Systems (EuroMLSys)}}.
\newblock


\bibitem[Brown et~al\mbox{.}(2020)]%
        {brown2020language}
\bibfield{author}{\bibinfo{person}{Tom Brown}, \bibinfo{person}{Benjamin Mann}, \bibinfo{person}{Nick Ryder}, \bibinfo{person}{Melanie Subbiah}, \bibinfo{person}{Jared~D Kaplan}, \bibinfo{person}{Prafulla Dhariwal}, \bibinfo{person}{Arvind Neelakantan}, \bibinfo{person}{Pranav Shyam}, \bibinfo{person}{Girish Sastry}, \bibinfo{person}{Amanda Askell}, {et~al\mbox{.}}} \bibinfo{year}{2020}\natexlab{}.
\newblock \showarticletitle{{Language Models are Few-Shot Learners}}.
\newblock \bibinfo{journal}{\emph{Advances in neural information processing systems}} (\bibinfo{year}{2020}).
\newblock


\bibitem[Cai et~al\mbox{.}(2024)]%
        {cai2024read}
\bibfield{author}{\bibinfo{person}{Ruisi Cai}, \bibinfo{person}{Yeonju Ro}, \bibinfo{person}{Geon-Woo Kim}, \bibinfo{person}{Peihao Wang}, \bibinfo{person}{Babak~Ehteshami Bejnordi}, \bibinfo{person}{Aditya Akella}, {and} \bibinfo{person}{Zhangyang Wang}.} \bibinfo{year}{2024}\natexlab{}.
\newblock \showarticletitle{{Read-ME: Refactorizing LLMs as Router-Decoupled Mixture of Experts with System Co-Design}}. In \bibinfo{booktitle}{\emph{The Thirty-eighth Annual Conference on Neural Information Processing Systems (NeurIPS)}}.
\newblock


\bibitem[Cohen et~al\mbox{.}(2009)]%
        {cohen2009pearson}
\bibfield{author}{\bibinfo{person}{Israel Cohen}, \bibinfo{person}{Yiteng Huang}, \bibinfo{person}{Jingdong Chen}, \bibinfo{person}{Jacob Benesty}, \bibinfo{person}{Jacob Benesty}, \bibinfo{person}{Jingdong Chen}, \bibinfo{person}{Yiteng Huang}, {and} \bibinfo{person}{Israel Cohen}.} \bibinfo{year}{2009}\natexlab{}.
\newblock \showarticletitle{{Pearson Correlation Coefficient}}.
\newblock \bibinfo{journal}{\emph{Noise Reduction in Speech Processing}} (\bibinfo{year}{2009}).
\newblock


\bibitem[Cormen et~al\mbox{.}(2022)]%
        {cormen2022introduction}
\bibfield{author}{\bibinfo{person}{Thomas~H Cormen}, \bibinfo{person}{Charles~E Leiserson}, \bibinfo{person}{Ronald~L Rivest}, {and} \bibinfo{person}{Clifford Stein}.} \bibinfo{year}{2022}\natexlab{}.
\newblock \bibinfo{booktitle}{\emph{{Introduction to Algorithms}}}.
\newblock \bibinfo{publisher}{MIT press}.
\newblock


\bibitem[Dai et~al\mbox{.}(2024a)]%
        {dai2024deepseekmoe}
\bibfield{author}{\bibinfo{person}{Damai Dai}, \bibinfo{person}{Chengqi Deng}, \bibinfo{person}{Chenggang Zhao}, \bibinfo{person}{RX Xu}, \bibinfo{person}{Huazuo Gao}, \bibinfo{person}{Deli Chen}, \bibinfo{person}{Jiashi Li}, \bibinfo{person}{Wangding Zeng}, \bibinfo{person}{Xingkai Yu}, \bibinfo{person}{Y Wu}, {et~al\mbox{.}}} \bibinfo{year}{2024}\natexlab{a}.
\newblock \showarticletitle{{DeepSeekMoE: Towards Ultimate Expert Specialization in Mixture-of-Experts Language Models}}.
\newblock \bibinfo{journal}{\emph{arXiv preprint arXiv:2401.06066}} (\bibinfo{year}{2024}).
\newblock


\bibitem[Dai et~al\mbox{.}(2024b)]%
        {dai2024neural}
\bibfield{author}{\bibinfo{person}{Sunhao Dai}, \bibinfo{person}{Yuqi Zhou}, \bibinfo{person}{Liang Pang}, \bibinfo{person}{Weihao Liu}, \bibinfo{person}{Xiaolin Hu}, \bibinfo{person}{Yong Liu}, \bibinfo{person}{Xiao Zhang}, \bibinfo{person}{Gang Wang}, {and} \bibinfo{person}{Jun Xu}.} \bibinfo{year}{2024}\natexlab{b}.
\newblock \showarticletitle{{Neural Retrievers are Biased Towards LLM-Generated Content}}. In \bibinfo{booktitle}{\emph{Proceedings of the 30th ACM SIGKDD Conference on Knowledge Discovery and Data Mining}}.
\newblock


\bibitem[{Dmitry Lepikhin, HyoukJoong Lee, Yuanzhong Xu, Dehao Chen, Orhan Firat, Yanping Huang, Maxim Krikun, Noam Shazeer, Zhifeng Chen}(2021)]%
        {kim2021moe}
\bibfield{author}{\bibinfo{person}{{Dmitry Lepikhin, HyoukJoong Lee, Yuanzhong Xu, Dehao Chen, Orhan Firat, Yanping Huang, Maxim Krikun, Noam Shazeer, Zhifeng Chen}}.} \bibinfo{year}{2021}\natexlab{}.
\newblock \showarticletitle{{GShard: Scaling Giant Models with Conditional Computation and Automatic Sharding}}. In \bibinfo{booktitle}{\emph{International Conference on Learning Representations (ICLR)}}.
\newblock


\bibitem[Du et~al\mbox{.}(2024)]%
        {du2024sida}
\bibfield{author}{\bibinfo{person}{Zhixu Du}, \bibinfo{person}{Shiyu Li}, \bibinfo{person}{Yuhao Wu}, \bibinfo{person}{Xiangyu Jiang}, \bibinfo{person}{Jingwei Sun}, \bibinfo{person}{Qilin Zheng}, \bibinfo{person}{Yongkai Wu}, \bibinfo{person}{Ang Li}, \bibinfo{person}{Hai Li}, {and} \bibinfo{person}{Yiran Chen}.} \bibinfo{year}{2024}\natexlab{}.
\newblock \showarticletitle{{SiDA: Sparsity-Inspired Data-Aware Serving for Efficient and Scalable Large Mixture-of-Experts Models}}.
\newblock \bibinfo{journal}{\emph{Proceedings of Machine Learning and Systems (MLSys)}} (\bibinfo{year}{2024}).
\newblock


\bibitem[Dumer(2007)]%
        {dumer2007covering}
\bibfield{author}{\bibinfo{person}{Ilya Dumer}.} \bibinfo{year}{2007}\natexlab{}.
\newblock \showarticletitle{{Covering Spheres with Spheres}}.
\newblock \bibinfo{journal}{\emph{Discrete \& Computational Geometry}} (\bibinfo{year}{2007}).
\newblock


\bibitem[Eliseev and Mazur(2023)]%
        {eliseev2023fast}
\bibfield{author}{\bibinfo{person}{Artyom Eliseev} {and} \bibinfo{person}{Denis Mazur}.} \bibinfo{year}{2023}\natexlab{}.
\newblock \showarticletitle{{Fast Inference of Mixture-of-Experts Language Models with Offloading}}.
\newblock \bibinfo{journal}{\emph{arXiv preprint arXiv:2312.17238}} (\bibinfo{year}{2023}).
\newblock


\bibitem[Elzinga and Hearn(1972)]%
        {elzinga1972minimum}
\bibfield{author}{\bibinfo{person}{D~Jack Elzinga} {and} \bibinfo{person}{Donald~W Hearn}.} \bibinfo{year}{1972}\natexlab{}.
\newblock \showarticletitle{{The Minimum Covering Sphere Problem}}.
\newblock \bibinfo{journal}{\emph{Management Science}} (\bibinfo{year}{1972}).
\newblock


\bibitem[Gupta et~al\mbox{.}(2024)]%
        {gupta2024lynx}
\bibfield{author}{\bibinfo{person}{Vima Gupta}, \bibinfo{person}{Kartik Sinha}, \bibinfo{person}{Ada Gavrilovska}, {and} \bibinfo{person}{Anand~Padmanabha Iyer}.} \bibinfo{year}{2024}\natexlab{}.
\newblock \showarticletitle{{Lynx: Enabling Efficient MoE Inference through Dynamic Batch-Aware Expert Selection}}.
\newblock \bibinfo{journal}{\emph{arXiv preprint arXiv:2411.08982}} (\bibinfo{year}{2024}).
\newblock


\bibitem[Harris et~al\mbox{.}(2020)]%
        {harris2020array}
\bibfield{author}{\bibinfo{person}{Charles~R. Harris}, \bibinfo{person}{K.~Jarrod Millman}, \bibinfo{person}{St{\'{e}}fan~J. van~der Walt}, \bibinfo{person}{Ralf Gommers}, \bibinfo{person}{Pauli Virtanen}, \bibinfo{person}{David Cournapeau}, \bibinfo{person}{Eric Wieser}, \bibinfo{person}{Julian Taylor}, \bibinfo{person}{Sebastian Berg}, \bibinfo{person}{Nathaniel~J. Smith}, \bibinfo{person}{Robert Kern}, \bibinfo{person}{Matti Picus}, \bibinfo{person}{Stephan Hoyer}, \bibinfo{person}{Marten~H. van Kerkwijk}, \bibinfo{person}{Matthew Brett}, \bibinfo{person}{Allan Haldane}, \bibinfo{person}{Jaime~Fern{\'{a}}ndez del R{\'{i}}o}, \bibinfo{person}{Mark Wiebe}, \bibinfo{person}{Pearu Peterson}, \bibinfo{person}{Pierre G{\'{e}}rard-Marchant}, \bibinfo{person}{Kevin Sheppard}, \bibinfo{person}{Tyler Reddy}, \bibinfo{person}{Warren Weckesser}, \bibinfo{person}{Hameer Abbasi}, \bibinfo{person}{Christoph Gohlke}, {and} \bibinfo{person}{Travis~E. Oliphant}.} \bibinfo{year}{2020}\natexlab{}.
\newblock \showarticletitle{{Array Programming with NumPy}}.
\newblock \bibinfo{journal}{\emph{Nature}} (\bibinfo{year}{2020}).
\newblock


\bibitem[He et~al\mbox{.}(2025)]%
        {he2025capacity}
\bibfield{author}{\bibinfo{person}{Shwai He}, \bibinfo{person}{Weilin Cai}, \bibinfo{person}{Jiayi Huang}, {and} \bibinfo{person}{Ang Li}.} \bibinfo{year}{2025}\natexlab{}.
\newblock \showarticletitle{{Capacity-Aware Inference: Mitigating the Straggler Effect in Mixture of Experts}}.
\newblock \bibinfo{journal}{\emph{arXiv preprint arXiv:2503.05066}} (\bibinfo{year}{2025}).
\newblock


\bibitem[He et~al\mbox{.}(2024)]%
        {he2024towards}
\bibfield{author}{\bibinfo{person}{Shwai He}, \bibinfo{person}{Daize Dong}, \bibinfo{person}{Liang Ding}, {and} \bibinfo{person}{Ang Li}.} \bibinfo{year}{2024}\natexlab{}.
\newblock \showarticletitle{{Towards Efficient Mixture of Experts: A Holistic Study of Compression Techniques}}.
\newblock \bibinfo{journal}{\emph{arXiv preprint arXiv:2406.02500}} (\bibinfo{year}{2024}).
\newblock


\bibitem[Hwang et~al\mbox{.}(2024)]%
        {hwang2024pre}
\bibfield{author}{\bibinfo{person}{Ranggi Hwang}, \bibinfo{person}{Jianyu Wei}, \bibinfo{person}{Shijie Cao}, \bibinfo{person}{Changho Hwang}, \bibinfo{person}{Xiaohu Tang}, \bibinfo{person}{Ting Cao}, {and} \bibinfo{person}{Mao Yang}.} \bibinfo{year}{2024}\natexlab{}.
\newblock \showarticletitle{{Pre-gated MoE: An Algorithm-System Co-Design for Fast and Scalable Mixture-of-Expert Inference}}. In \bibinfo{booktitle}{\emph{2024 ACM/IEEE 51st Annual International Symposium on Computer Architecture (ISCA)}}.
\newblock


\bibitem[Jiang et~al\mbox{.}(2024b)]%
        {jiang2024mixtral}
\bibfield{author}{\bibinfo{person}{Albert~Q Jiang}, \bibinfo{person}{Alexandre Sablayrolles}, \bibinfo{person}{Antoine Roux}, \bibinfo{person}{Arthur Mensch}, \bibinfo{person}{Blanche Savary}, \bibinfo{person}{Chris Bamford}, \bibinfo{person}{Devendra~Singh Chaplot}, \bibinfo{person}{Diego de~las Casas}, \bibinfo{person}{Emma~Bou Hanna}, \bibinfo{person}{Florian Bressand}, {et~al\mbox{.}}} \bibinfo{year}{2024}\natexlab{b}.
\newblock \showarticletitle{{Mixtral of Experts}}.
\newblock \bibinfo{journal}{\emph{arXiv preprint arXiv:2401.04088}} (\bibinfo{year}{2024}).
\newblock


\bibitem[Jiang et~al\mbox{.}(2024a)]%
        {jiang2024lilac}
\bibfield{author}{\bibinfo{person}{Zhihan Jiang}, \bibinfo{person}{Jinyang Liu}, \bibinfo{person}{Zhuangbin Chen}, \bibinfo{person}{Yichen Li}, \bibinfo{person}{Junjie Huang}, \bibinfo{person}{Yintong Huo}, \bibinfo{person}{Pinjia He}, \bibinfo{person}{Jiazhen Gu}, {and} \bibinfo{person}{Michael~R Lyu}.} \bibinfo{year}{2024}\natexlab{a}.
\newblock \showarticletitle{{LILAC: Log Parsing using LLMs with Adaptive Parsing Cache}}.
\newblock \bibinfo{journal}{\emph{Proceedings of the ACM on Software Engineering}} (\bibinfo{year}{2024}).
\newblock


\bibitem[Jie et~al\mbox{.}(2025)]%
        {jie2025mixture}
\bibfield{author}{\bibinfo{person}{Shibo Jie}, \bibinfo{person}{Yehui Tang}, \bibinfo{person}{Kai Han}, \bibinfo{person}{Yitong Li}, \bibinfo{person}{Duyu Tang}, \bibinfo{person}{Zhi-Hong Deng}, {and} \bibinfo{person}{Yunhe Wang}.} \bibinfo{year}{2025}\natexlab{}.
\newblock \showarticletitle{{Mixture of Lookup Experts}}. In \bibinfo{booktitle}{\emph{International Conference on Machine Learning (ICML)}}.
\newblock


\bibitem[Kamahori et~al\mbox{.}(2025)]%
        {kamahori2025fiddler}
\bibfield{author}{\bibinfo{person}{Keisuke Kamahori}, \bibinfo{person}{Tian Tang}, \bibinfo{person}{Yile Gu}, \bibinfo{person}{Kan Zhu}, {and} \bibinfo{person}{Baris Kasikci}.} \bibinfo{year}{2025}\natexlab{}.
\newblock \showarticletitle{{Fiddler: CPU-GPU Orchestration for Fast Inference of Mixture-of-Experts Models}}. In \bibinfo{booktitle}{\emph{International Conference on Learning Representations (ICLR)}}.
\newblock


\bibitem[Kim et~al\mbox{.}(2023)]%
        {kim2023mixture}
\bibfield{author}{\bibinfo{person}{Young~Jin Kim}, \bibinfo{person}{Raffy Fahim}, {and} \bibinfo{person}{Hany~Hassan Awadalla}.} \bibinfo{year}{2023}\natexlab{}.
\newblock \showarticletitle{{Mixture of Quantized Experts (MoQE): Complementary Effect of Low-bit Quantization and Robustness}}.
\newblock \bibinfo{journal}{\emph{arXiv preprint arXiv:2310.02410}} (\bibinfo{year}{2023}).
\newblock


\bibitem[Kwon et~al\mbox{.}(2023)]%
        {kwon2023efficient}
\bibfield{author}{\bibinfo{person}{Woosuk Kwon}, \bibinfo{person}{Zhuohan Li}, \bibinfo{person}{Siyuan Zhuang}, \bibinfo{person}{Ying Sheng}, \bibinfo{person}{Lianmin Zheng}, \bibinfo{person}{Cody~Hao Yu}, \bibinfo{person}{Joseph Gonzalez}, \bibinfo{person}{Hao Zhang}, {and} \bibinfo{person}{Ion Stoica}.} \bibinfo{year}{2023}\natexlab{}.
\newblock \showarticletitle{{Efficient Memory Management for Large Language Model Serving with PagedAttention}}. In \bibinfo{booktitle}{\emph{Proceedings of the 29th Symposium on Operating Systems Principles (SOSP)}}.
\newblock


\bibitem[Lee et~al\mbox{.}(2012)]%
        {lee2012prefetching}
\bibfield{author}{\bibinfo{person}{Jaekyu Lee}, \bibinfo{person}{Hyesoon Kim}, {and} \bibinfo{person}{Richard Vuduc}.} \bibinfo{year}{2012}\natexlab{}.
\newblock \showarticletitle{{When Prefetching Works, When It Doesn’t, and Why}}.
\newblock \bibinfo{journal}{\emph{ACM Transactions on Architecture and Code Optimization (TACO)}} (\bibinfo{year}{2012}).
\newblock


\bibitem[Lee et~al\mbox{.}(2024b)]%
        {lee2024stun}
\bibfield{author}{\bibinfo{person}{Jaeseong Lee}, \bibinfo{person}{Aurick Qiao}, \bibinfo{person}{Daniel~F Campos}, \bibinfo{person}{Zhewei Yao}, \bibinfo{person}{Yuxiong He}, {et~al\mbox{.}}} \bibinfo{year}{2024}\natexlab{b}.
\newblock \showarticletitle{{STUN: Structured-Then-Unstructured Pruning for Scalable MoE Pruning}}.
\newblock \bibinfo{journal}{\emph{arXiv preprint arXiv:2409.06211}} (\bibinfo{year}{2024}).
\newblock


\bibitem[Lee et~al\mbox{.}(2024a)]%
        {lee2024infinigen}
\bibfield{author}{\bibinfo{person}{Wonbeom Lee}, \bibinfo{person}{Jungi Lee}, \bibinfo{person}{Junghwan Seo}, {and} \bibinfo{person}{Jaewoong Sim}.} \bibinfo{year}{2024}\natexlab{a}.
\newblock \showarticletitle{{InfiniGen: Efficient Generative Inference of Large Language Models with Dynamic KV Cache Management}}. In \bibinfo{booktitle}{\emph{18th USENIX Symposium on Operating Systems Design and Implementation (OSDI)}}.
\newblock


\bibitem[Li et~al\mbox{.}(2023)]%
        {li2023accelerating}
\bibfield{author}{\bibinfo{person}{Jiamin Li}, \bibinfo{person}{Yimin Jiang}, \bibinfo{person}{Yibo Zhu}, \bibinfo{person}{Cong Wang}, {and} \bibinfo{person}{Hong Xu}.} \bibinfo{year}{2023}\natexlab{}.
\newblock \showarticletitle{{Accelerating Distributed MoE training and inference with Lina}}. In \bibinfo{booktitle}{\emph{2023 USENIX Annual Technical Conference (USENIX ATC 23)}}.
\newblock


\bibitem[Li et~al\mbox{.}(2024)]%
        {li2024go}
\bibfield{author}{\bibinfo{person}{Yichen Li}, \bibinfo{person}{Yintong Huo}, \bibinfo{person}{Renyi Zhong}, \bibinfo{person}{Zhihan Jiang}, \bibinfo{person}{Jinyang Liu}, \bibinfo{person}{Junjie Huang}, \bibinfo{person}{Jiazhen Gu}, \bibinfo{person}{Pinjia He}, {and} \bibinfo{person}{Michael~R Lyu}.} \bibinfo{year}{2024}\natexlab{}.
\newblock \showarticletitle{{Go Static: Contextualized Logging Statement Generation}}.
\newblock \bibinfo{journal}{\emph{Proceedings of the ACM on Software Engineering}} (\bibinfo{year}{2024}).
\newblock


\bibitem[Lin et~al\mbox{.}(2024)]%
        {lin2024data}
\bibfield{author}{\bibinfo{person}{Xinyu Lin}, \bibinfo{person}{Wenjie Wang}, \bibinfo{person}{Yongqi Li}, \bibinfo{person}{Shuo Yang}, \bibinfo{person}{Fuli Feng}, \bibinfo{person}{Yinwei Wei}, {and} \bibinfo{person}{Tat-Seng Chua}.} \bibinfo{year}{2024}\natexlab{}.
\newblock \showarticletitle{{Data-efficient Fine-tuning for LLM-based Recommendation}}. In \bibinfo{booktitle}{\emph{Proceedings of the 47th International ACM SIGIR Conference on Research and Development in Information Retrieval}}.
\newblock


\bibitem[Liu et~al\mbox{.}(2024a)]%
        {liu2024deepseek}
\bibfield{author}{\bibinfo{person}{Aixin Liu}, \bibinfo{person}{Bei Feng}, \bibinfo{person}{Bing Xue}, \bibinfo{person}{Bingxuan Wang}, \bibinfo{person}{Bochao Wu}, \bibinfo{person}{Chengda Lu}, \bibinfo{person}{Chenggang Zhao}, \bibinfo{person}{Chengqi Deng}, \bibinfo{person}{Chenyu Zhang}, \bibinfo{person}{Chong Ruan}, {et~al\mbox{.}}} \bibinfo{year}{2024}\natexlab{a}.
\newblock \showarticletitle{{DeepSeek-V3 Technical Report}}.
\newblock \bibinfo{journal}{\emph{arXiv preprint arXiv:2412.19437}} (\bibinfo{year}{2024}).
\newblock


\bibitem[Liu et~al\mbox{.}(2025)]%
        {liu2025optimizing}
\bibfield{author}{\bibinfo{person}{Mengfan Liu}, \bibinfo{person}{Wei Wang}, {and} \bibinfo{person}{Chuan Wu}.} \bibinfo{year}{2025}\natexlab{}.
\newblock \showarticletitle{Optimizing Distributed Deployment of Mixture-of-Experts Model Inference in Serverless Computing}. In \bibinfo{booktitle}{\emph{IEEE Conference on Computer Communications (INFOCOM)}}.
\newblock


\bibitem[Liu et~al\mbox{.}(2024b)]%
        {liu2024cachegen}
\bibfield{author}{\bibinfo{person}{Yuhan Liu}, \bibinfo{person}{Hanchen Li}, \bibinfo{person}{Yihua Cheng}, \bibinfo{person}{Siddhant Ray}, \bibinfo{person}{Yuyang Huang}, \bibinfo{person}{Qizheng Zhang}, \bibinfo{person}{Kuntai Du}, \bibinfo{person}{Jiayi Yao}, \bibinfo{person}{Shan Lu}, \bibinfo{person}{Ganesh Ananthanarayanan}, {et~al\mbox{.}}} \bibinfo{year}{2024}\natexlab{b}.
\newblock \showarticletitle{{CacheGen: KV Cache Compression and Streaming for Fast Large Language Model Serving}}. In \bibinfo{booktitle}{\emph{Proceedings of the ACM SIGCOMM 2024 Conference}}.
\newblock


\bibitem[Mikolov et~al\mbox{.}(2013)]%
        {mikolov2013efficient}
\bibfield{author}{\bibinfo{person}{Tomas Mikolov}, \bibinfo{person}{Kai Chen}, \bibinfo{person}{Greg Corrado}, {and} \bibinfo{person}{Jeffrey Dean}.} \bibinfo{year}{2013}\natexlab{}.
\newblock \showarticletitle{{Efficient Estimation of Word Representations in Vector Space}}.
\newblock  (\bibinfo{year}{2013}).
\newblock


\bibitem[Nam et~al\mbox{.}(2024)]%
        {nam2024using}
\bibfield{author}{\bibinfo{person}{Daye Nam}, \bibinfo{person}{Andrew Macvean}, \bibinfo{person}{Vincent Hellendoorn}, \bibinfo{person}{Bogdan Vasilescu}, {and} \bibinfo{person}{Brad Myers}.} \bibinfo{year}{2024}\natexlab{}.
\newblock \showarticletitle{{Using an LLM to Help with Code Understanding}}. In \bibinfo{booktitle}{\emph{Proceedings of the IEEE/ACM 46th International Conference on Software Engineering}}.
\newblock


\bibitem[{NVIDIA}(2024)]%
        {cuda-runtime-api}
\bibfield{author}{\bibinfo{person}{{NVIDIA}}.} \bibinfo{year}{2024}\natexlab{}.
\newblock \bibinfo{title}{{CUDA Runtime API :: CUDA Toolkit Documentation}}.
\newblock \bibinfo{howpublished}{\url{https://docs.nvidia.com/cuda/cuda-runtime-api/index.html}}.
\newblock


\bibitem[{Ollama}(2024)]%
        {ollama}
\bibfield{author}{\bibinfo{person}{{Ollama}}.} \bibinfo{year}{2024}\natexlab{}.
\newblock \bibinfo{title}{{Get Up and Running with Large Language Models.}}
\newblock \bibinfo{howpublished}{\url{https://ollama.com/}}.
\newblock


\bibitem[Paszke et~al\mbox{.}(2019)]%
        {paszke2019pytorch}
\bibfield{author}{\bibinfo{person}{Adam Paszke}, \bibinfo{person}{Sam Gross}, \bibinfo{person}{Francisco Massa}, \bibinfo{person}{Adam Lerer}, \bibinfo{person}{James Bradbury}, \bibinfo{person}{Gregory Chanan}, \bibinfo{person}{Trevor Killeen}, \bibinfo{person}{Zeming Lin}, \bibinfo{person}{Natalia Gimelshein}, \bibinfo{person}{Luca Antiga}, {et~al\mbox{.}}} \bibinfo{year}{2019}\natexlab{}.
\newblock \showarticletitle{{PyTorch: An Imperative Style, High-Performance Deep Learning Library}}.
\newblock \bibinfo{journal}{\emph{Advances in Neural Information Processing Systems (NIPS)}} (\bibinfo{year}{2019}).
\newblock


\bibitem[Patel et~al\mbox{.}(2024)]%
        {patel2024splitwise}
\bibfield{author}{\bibinfo{person}{Pratyush Patel}, \bibinfo{person}{Esha Choukse}, \bibinfo{person}{Chaojie Zhang}, \bibinfo{person}{Aashaka Shah}, \bibinfo{person}{{\'I}{\~n}igo Goiri}, \bibinfo{person}{Saeed Maleki}, {and} \bibinfo{person}{Ricardo Bianchini}.} \bibinfo{year}{2024}\natexlab{}.
\newblock \showarticletitle{{Splitwise: Efficient Generative LLM Inference Using Phase Splitting}}. In \bibinfo{booktitle}{\emph{2024 ACM/IEEE 51st Annual International Symposium on Computer Architecture (ISCA)}}.
\newblock


\bibitem[{Pingzhi Li, Zhenyu Zhang, Prateek Yadav, Yi-Lin Sung, Yu Cheng, Mohit Bansal, Tianlong Chen}(2024)]%
        {li2023merge}
\bibfield{author}{\bibinfo{person}{{Pingzhi Li, Zhenyu Zhang, Prateek Yadav, Yi-Lin Sung, Yu Cheng, Mohit Bansal, Tianlong Chen}}.} \bibinfo{year}{2024}\natexlab{}.
\newblock \showarticletitle{{Merge, Then Compress: Demystify Efficient SMoE with Hints from Its Routing Policy}}. In \bibinfo{booktitle}{\emph{International Conference on Learning Representations (ICLR)}}.
\newblock


\bibitem[Radford et~al\mbox{.}(2019)]%
        {radford2019language}
\bibfield{author}{\bibinfo{person}{Alec Radford}, \bibinfo{person}{Jeffrey Wu}, \bibinfo{person}{Rewon Child}, \bibinfo{person}{David Luan}, \bibinfo{person}{Dario Amodei}, \bibinfo{person}{Ilya Sutskever}, {et~al\mbox{.}}} \bibinfo{year}{2019}\natexlab{}.
\newblock \showarticletitle{{Language Models are Unsupervised Multitask Learners}}.
\newblock \bibinfo{journal}{\emph{OpenAI blog}} (\bibinfo{year}{2019}).
\newblock


\bibitem[Rankin(1947)]%
        {rankin1947closest}
\bibfield{author}{\bibinfo{person}{Robert~Alexander Rankin}.} \bibinfo{year}{1947}\natexlab{}.
\newblock \showarticletitle{{On the Closest Packing of Spheres in N Dimensions}}.
\newblock \bibinfo{journal}{\emph{Annals of Mathematics}} (\bibinfo{year}{1947}).
\newblock


\bibitem[{Rui Kong, Yuanchun Li, Qingtian Feng, Weijun Wang, Xiaozhou Ye, Ye Ouyang, Linghe Kong, Yunxin Liu}(2023)]%
        {swapmoe}
\bibfield{author}{\bibinfo{person}{{Rui Kong, Yuanchun Li, Qingtian Feng, Weijun Wang, Xiaozhou Ye, Ye Ouyang, Linghe Kong, Yunxin Liu}}.} \bibinfo{year}{2023}\natexlab{}.
\newblock \showarticletitle{{SwapMoE: Serving Off-the-shelf MoE-based Large Language Models with Tunable Memory Budget}}.
\newblock \bibinfo{journal}{\emph{arXiv preprint arXiv:2308.15030}} (\bibinfo{year}{2023}).
\newblock


\bibitem[Shannon(1948)]%
        {shannon1948mathematical}
\bibfield{author}{\bibinfo{person}{Claude~Elwood Shannon}.} \bibinfo{year}{1948}\natexlab{}.
\newblock \showarticletitle{A Mathematical Theory of Communication}.
\newblock \bibinfo{journal}{\emph{The Bell System Technical Journal}} (\bibinfo{year}{1948}).
\newblock


\bibitem[{ShareGPT}(2022)]%
        {sharegpt}
\bibfield{author}{\bibinfo{person}{{ShareGPT}}.} \bibinfo{year}{2022}\natexlab{}.
\newblock \bibinfo{title}{{ShareGPT: Share Your Wildest ChatGPT Conversations}}.
\newblock \bibinfo{howpublished}{\url{https://sharegpt.com/}}.
\newblock


\bibitem[{Snowflake}(2024)]%
        {snowflake-arctic}
\bibfield{author}{\bibinfo{person}{{Snowflake}}.} \bibinfo{year}{2024}\natexlab{}.
\newblock \bibinfo{title}{{Snowflake Arctic: The Best LLM for Enterprise AI}}.
\newblock \bibinfo{howpublished}{\url{https://www.snowflake.com/en/data-cloud/arctic/}}.
\newblock


\bibitem[Song et~al\mbox{.}(2024)]%
        {song2024promoe}
\bibfield{author}{\bibinfo{person}{Xiaoniu Song}, \bibinfo{person}{Zihang Zhong}, {and} \bibinfo{person}{Rong Chen}.} \bibinfo{year}{2024}\natexlab{}.
\newblock \showarticletitle{{ProMoE: Fast MoE-based LLM Serving using Proactive Caching}}.
\newblock \bibinfo{journal}{\emph{arXiv preprint arXiv:2410.22134}} (\bibinfo{year}{2024}).
\newblock


\bibitem[Stojkovic et~al\mbox{.}(2025)]%
        {stojkovic2025dynamollm}
\bibfield{author}{\bibinfo{person}{Jovan Stojkovic}, \bibinfo{person}{Chaojie Zhang}, \bibinfo{person}{{\'I}{\~n}igo Goiri}, \bibinfo{person}{Josep Torrellas}, {and} \bibinfo{person}{Esha Choukse}.} \bibinfo{year}{2025}\natexlab{}.
\newblock \showarticletitle{{DynamoLLM: Designing LLM Inference Clusters for Performance and Energy Efficiency}}. In \bibinfo{booktitle}{\emph{International Symposium on High-Performance Computer Architecture (HPCA)}}.
\newblock


\bibitem[Tang et~al\mbox{.}(2024)]%
        {tang2024hobbit}
\bibfield{author}{\bibinfo{person}{Peng Tang}, \bibinfo{person}{Jiacheng Liu}, \bibinfo{person}{Xiaofeng Hou}, \bibinfo{person}{Yifei Pu}, \bibinfo{person}{Jing Wang}, \bibinfo{person}{Pheng-Ann Heng}, \bibinfo{person}{Chao Li}, {and} \bibinfo{person}{Minyi Guo}.} \bibinfo{year}{2024}\natexlab{}.
\newblock \showarticletitle{Hobbit: A Mixed Precision Expert Offloading System for Fast MoE Inference}.
\newblock \bibinfo{journal}{\emph{arXiv preprint arXiv:2411.01433}} (\bibinfo{year}{2024}).
\newblock


\bibitem[Vaswani(2017)]%
        {vaswani2017attention}
\bibfield{author}{\bibinfo{person}{A Vaswani}.} \bibinfo{year}{2017}\natexlab{}.
\newblock \showarticletitle{Attention is all you need}.
\newblock \bibinfo{journal}{\emph{Advances in Neural Information Processing Systems}} (\bibinfo{year}{2017}).
\newblock


\bibitem[Wolf et~al\mbox{.}(2020)]%
        {wolf2020huggingface}
\bibfield{author}{\bibinfo{person}{Thomas Wolf}, \bibinfo{person}{Lysandre Debut}, \bibinfo{person}{Victor Sanh}, \bibinfo{person}{Julien Chaumond}, \bibinfo{person}{Clement Delangue}, \bibinfo{person}{Anthony Moi}, \bibinfo{person}{Pierric Cistac}, \bibinfo{person}{Tim Rault}, \bibinfo{person}{Remi Louf}, \bibinfo{person}{Morgan Funtowicz}, \bibinfo{person}{Joe Davison}, \bibinfo{person}{Sam Shleifer}, \bibinfo{person}{Patrick von Platen}, \bibinfo{person}{Clara Ma}, \bibinfo{person}{Yacine Jernite}, \bibinfo{person}{Julien Plu}, \bibinfo{person}{Canwen Xu}, \bibinfo{person}{Teven Le~Scao}, \bibinfo{person}{Sylvain Gugger}, \bibinfo{person}{Mariama Drame}, \bibinfo{person}{Quentin Lhoest}, {and} \bibinfo{person}{Alexander Rush}.} \bibinfo{year}{2020}\natexlab{}.
\newblock \showarticletitle{{HuggingFace's Transformers: State-of-the-Art Natural Language Processing}}. In \bibinfo{booktitle}{\emph{Proceedings of the 2020 Conference on Empirical Methods in Natural Language Processing: System Demonstrations}}.
\newblock


\bibitem[Wu et~al\mbox{.}(2025)]%
        {wu2025samoyeds}
\bibfield{author}{\bibinfo{person}{Chenpeng Wu}, \bibinfo{person}{Qiqi Gu}, \bibinfo{person}{Heng Shi}, \bibinfo{person}{Jianguo Yao}, {and} \bibinfo{person}{Haibing Guan}.} \bibinfo{year}{2025}\natexlab{}.
\newblock \showarticletitle{{Samoyeds: Accelerating MoE Models with Structured Sparsity Leveraging Sparse Tensor Cores}}. In \bibinfo{booktitle}{\emph{Proceedings of the Twentieth European Conference on Computer Systems (EuroSys)}}.
\newblock


\bibitem[{xAI}(2023)]%
        {xai-grok}
\bibfield{author}{\bibinfo{person}{{xAI}}.} \bibinfo{year}{2023}\natexlab{}.
\newblock \bibinfo{title}{{Announcing Grok}}.
\newblock \bibinfo{howpublished}{\url{https://x.ai/blog/grok}}.
\newblock


\bibitem[Xue et~al\mbox{.}(2024)]%
        {xue2024moe}
\bibfield{author}{\bibinfo{person}{Leyang Xue}, \bibinfo{person}{Yao Fu}, \bibinfo{person}{Zhan Lu}, \bibinfo{person}{Luo Mai}, {and} \bibinfo{person}{Mahesh Marina}.} \bibinfo{year}{2024}\natexlab{}.
\newblock \showarticletitle{{MoE-Infinity: Efficient MoE Inference on Personal Machines with Sparsity-Aware Expert Cache}}.
\newblock \bibinfo{journal}{\emph{arXiv preprint arXiv:2401.14361}} (\bibinfo{year}{2024}).
\newblock


\bibitem[{Xue, Leyang and Fu, Yao and Lu, Zhan and Mai, Luo and Marina, Mahesh}({[n.\,d.]})]%
        {moe-infinity-code}
\bibfield{author}{\bibinfo{person}{{Xue, Leyang and Fu, Yao and Lu, Zhan and Mai, Luo and Marina, Mahesh}}.} \bibinfo{year}{[n.\,d.]}\natexlab{}.
\newblock \bibinfo{title}{{MoE-Infinity Codebase}}.
\newblock \bibinfo{howpublished}{\url{https://github.com/TorchMoE/MoE-Infinity}}.
\newblock


\bibitem[Yang et~al\mbox{.}(2024)]%
        {yang2024qwen2}
\bibfield{author}{\bibinfo{person}{An Yang}, \bibinfo{person}{Baosong Yang}, \bibinfo{person}{Binyuan Hui}, \bibinfo{person}{Bo Zheng}, \bibinfo{person}{Bowen Yu}, \bibinfo{person}{Chang Zhou}, \bibinfo{person}{Chengpeng Li}, \bibinfo{person}{Chengyuan Li}, \bibinfo{person}{Dayiheng Liu}, \bibinfo{person}{Fei Huang}, {et~al\mbox{.}}} \bibinfo{year}{2024}\natexlab{}.
\newblock \showarticletitle{{Qwen2 Technical Report}}.
\newblock \bibinfo{journal}{\emph{arXiv preprint arXiv:2407.10671}} (\bibinfo{year}{2024}).
\newblock


\bibitem[Yuksel et~al\mbox{.}(2012)]%
        {yuksel2012twenty}
\bibfield{author}{\bibinfo{person}{Seniha~Esen Yuksel}, \bibinfo{person}{Joseph~N Wilson}, {and} \bibinfo{person}{Paul~D Gader}.} \bibinfo{year}{2012}\natexlab{}.
\newblock \showarticletitle{{Twenty Years of Mixture of Experts}}.
\newblock \bibinfo{journal}{\emph{IEEE Transactions on Neural Networks and Learning Systems (TNNLS)}} (\bibinfo{year}{2012}).
\newblock


\bibitem[Zhang et~al\mbox{.}(2025)]%
        {zhang2025daop}
\bibfield{author}{\bibinfo{person}{Yujie Zhang}, \bibinfo{person}{Shivam Aggarwal}, {and} \bibinfo{person}{Tulika Mitra}.} \bibinfo{year}{2025}\natexlab{}.
\newblock \showarticletitle{{DAOP: Data-Aware Offloading and Predictive Pre-Calculation for Efficient MoE Inference}}. In \bibinfo{booktitle}{\emph{Design Automation and Test in Europe (DATE)}}.
\newblock


\bibitem[Zhao et~al\mbox{.}(2024)]%
        {zhao2024let}
\bibfield{author}{\bibinfo{person}{Yuyue Zhao}, \bibinfo{person}{Jiancan Wu}, \bibinfo{person}{Xiang Wang}, \bibinfo{person}{Wei Tang}, \bibinfo{person}{Dingxian Wang}, {and} \bibinfo{person}{Maarten de Rijke}.} \bibinfo{year}{2024}\natexlab{}.
\newblock \showarticletitle{{Let Me Do It for You: Towards LLM Empowered Recommendation via Tool Learning}}. In \bibinfo{booktitle}{\emph{Proceedings of the 47th International ACM SIGIR Conference on Research and Development in Information Retrieval}}.
\newblock


\bibitem[Zheng et~al\mbox{.}(2023)]%
        {zheng2023lmsys}
\bibfield{author}{\bibinfo{person}{Lianmin Zheng}, \bibinfo{person}{Wei-Lin Chiang}, \bibinfo{person}{Ying Sheng}, \bibinfo{person}{Tianle Li}, \bibinfo{person}{Siyuan Zhuang}, \bibinfo{person}{Zhanghao Wu}, \bibinfo{person}{Yonghao Zhuang}, \bibinfo{person}{Zhuohan Li}, \bibinfo{person}{Zi Lin}, \bibinfo{person}{Eric~P Xing}, {et~al\mbox{.}}} \bibinfo{year}{2023}\natexlab{}.
\newblock \showarticletitle{{LMSYS-Chat-1M: A Large-Scale Real-World LLM Conversation Dataset}}.
\newblock \bibinfo{journal}{\emph{arXiv preprint arXiv:2309.11998}} (\bibinfo{year}{2023}).
\newblock


\bibitem[Zhong et~al\mbox{.}(2024)]%
        {zhong2024distserve}
\bibfield{author}{\bibinfo{person}{Yinmin Zhong}, \bibinfo{person}{Shengyu Liu}, \bibinfo{person}{Junda Chen}, \bibinfo{person}{Jianbo Hu}, \bibinfo{person}{Yibo Zhu}, \bibinfo{person}{Xuanzhe Liu}, \bibinfo{person}{Xin Jin}, {and} \bibinfo{person}{Hao Zhang}.} \bibinfo{year}{2024}\natexlab{}.
\newblock \showarticletitle{DistServe: Disaggregating Prefill and Decoding for Goodput-optimized Large Language Model Serving}. In \bibinfo{booktitle}{\emph{18th USENIX Symposium on Operating Systems Design and Implementation (OSDI)}}.
\newblock


\bibitem[Zhou et~al\mbox{.}(2025)]%
        {zhou2025floe}
\bibfield{author}{\bibinfo{person}{Yuxin Zhou}, \bibinfo{person}{Zheng Li}, \bibinfo{person}{Jun Zhang}, \bibinfo{person}{Jue Wang}, \bibinfo{person}{Yiping Wang}, \bibinfo{person}{Zhongle Xie}, \bibinfo{person}{Ke Chen}, {and} \bibinfo{person}{Lidan Shou}.} \bibinfo{year}{2025}\natexlab{}.
\newblock \showarticletitle{{FloE: On-the-Fly MoE Inference on Memory-constrained GPU}}. In \bibinfo{booktitle}{\emph{International Conference on Machine Learning (ICML)}}.
\newblock


\end{thebibliography}
